\documentclass[preprint,review,12pt,number,sort&compress]{elsarticle}

\usepackage{amssymb}
\usepackage{colortbl}
\usepackage{bm,makecell}
\usepackage{amsmath,algorithm,algpseudocode,multirow,algorithmicx,subfigure}
\usepackage{booktabs,soul,xcolor}
\newsavebox\CBox
\def\textBF#1{\sbox\CBox{#1}\resizebox{\wd\CBox}{\ht\CBox}{\textbf{#1}}}
\usepackage{array}
\sethlcolor{yellow}
\soulregister\cite7
\soulregister\ref7 
\soulregister\citep7 
\soulregister\citet7 
\soulregister\pageref7 

\usepackage{lipsum}
\usepackage{framed}
\usepackage{enumitem}

\newcommand\ie{\textit{i.e.}}
\newcommand\eg{\textit{e.g.}}

\DeclareMathOperator{\argmin}{argmin}
\DeclareMathOperator{\argmax}{argmax}

\algdef{SE}[DOWHILE]{Do}{doWhile}{\algorithmicdo}[1]{\algorithmicwhile\ #1}%
\usepackage{jabbrv}

\journal{Pattern Recognition}

\begin{document}

\begin{frontmatter}



\title{A Max-relevance-min-divergence Criterion for Data Discretization with Applications on Naive Bayes}


\author[ad1]{Shihe Wang}     
\author[ad1]{Jianfeng Ren\corref{1}}
\cortext[1]{Corresponding author: Jianfeng.Ren@nottingham.edu.cn}        

\author[ad1]{Ruibin Bai}
\author[ad1]{Yuan Yao}
\author[ad2]{Xudong Jiang}
\address[ad1]{School of Computer Science, University of Nottingham Ningbo China, Ningbo 315100, China}
\address[ad2]{School of Electrical and Electronic Engineering, Nanyang Technological University, 639798, Singapore}
\begin{abstract}
In many classification models, data is discretized to better estimate its distribution. 
Existing discretization methods often target at
maximizing the discriminant power of discretized data, while overlooking the fact that the primary target of data discretization in classification is to improve the generalization performance. As a result, the data tend to be over-split into many small bins since the data without discretization retain the maximal discriminant information.
Thus, we propose a Max-Dependency-Min-Divergence (MDmD) criterion that maximizes both the discriminant information and generalization ability of the discretized data. 
More specifically, the Max-Dependency criterion maximizes the statistical dependency between the discretized data and the classification variable while the 
Min-Divergence criterion explicitly minimizes the JS-divergence between the training data and the validation data for a given discretization scheme.
The proposed MDmD criterion is technically appealing, but it is difficult to reliably estimate the high-order joint distributions of attributes and the classification variable. We hence further propose a more practical solution, Max-Relevance-Min-Divergence (MRmD) discretization scheme, where each attribute is discretized separately, by simultaneously maximizing the discriminant information and the generalization ability of the discretized data.
The proposed MRmD is compared with the state-of-the-art discretization algorithms under the naive Bayes classification framework on 45 benchmark datasets. It significantly outperforms all the compared methods on most of the datasets.
 
\end{abstract}

\begin{keyword}
Data Discretization, Maximal Dependency, Maximal Relevance, Minimal Divergence, Naive Bayes Classification
\end{keyword}

\end{frontmatter}

\section{{Introduction}}

{Deep-learning models have been successful in many applications \cite{gorishniy2021revisiting}, but they require a large amount of training samples. For applications such as drug discovery \cite{vamathevan2019applications} and medical diagnosis \cite{shaban2021accurate}, it is labor-expensive to collect many samples, where traditional machine-learning methods with much fewer model parameters may generalize better, \eg, decision tree~\cite{wang2020linear}, fuzzy rule-based classifiers~\cite{mu2020parallel}, naive Bayes~\cite{RNB2020shihe,zhang2021attribute,jiang2019class}, {k-nearest-neighbor classifier \cite{tran2017new}, and support vector machine \cite{rahman2016discretization,ramirez2015multivariate}}. To improve the generalization capability, and to handle mixed-type data,} 
continuous attributes are often discretized to facilitate a better estimation of the data distribution for subsequent classifiers~\cite{tahan2018emdid,tsai2019optimal,sharmin2019simultaneous,rahman2016discretization}. 
The discrete features are easier to understand than continuous ones because they are closer to knowledge-level representation~\cite{ramirez2015multivariate}.
Most importantly, by discretizing similar values into one bin, {the distribution discrepancy between training data and test data could be reduced, and hence} the generalization capability of a classifier could be enhanced~\cite{RNB2020shihe,zhang2021attribute,jiang2019class}. 
Data discretization aims to find a minimal set of cut points that optimally discretize continuous attributes to maximize the classification accuracy~\cite{tahan2018emdid,ramirez2015multivariate}. Existing methods often over-emphasize maximizing the discriminant information, while neglecting the primary target of data discretization, \ie, to improve the generalization performance by reducing the noisy information that is harmful to reliable classification. In data discretization, two opposing goals often compete with each other, \ie, the generalization performance is maximized when all samples are discretized into one bin so that there is no distribution discrepancy between training data and test data, but the discriminant information is totally lost in this case. On the other hand, the discriminant information is maximized when no discretization is performed on the data, but the generalization performance would not be improved. 

In literature, many discretization algorithms have been developed to maximize the dependence between discrete attributes and classification variables in terms of mutual information \cite{sharmin2019simultaneous}, information entropy \cite{xun2021novel,fayyad1993multi}, contingency coefficient \cite{tsai2008discretization,gonzalez2009ameva}, statistical interdependency \cite{cano2016laim,kurgan2004caim}, and many others~\cite{tay2002modified,ramirez2015multivariate,tahan2018emdid,zhou2021evolutionary}. 
However, none of them explicitly maximizes the generalization capability. Instead, they often restrict the number of intervals after discretization to be small, in the hope of retaining the generalization ability, \eg, Class-Attribute Interdependence Maximization (CAIM)~\cite{kurgan2004caim} and Class-Attribute Contingency Coefficient (CACC)~\cite{tsai2008discretization} both restrict the number of intervals to the number of classes. Such a design does not optimize the discretization scheme in terms of the generalization.

To tackle this problem, a Max-Dependency-Min-Divergence (MDmD) criterion is proposed to simultaneously maximize the discriminant power and the generalization ability. 
The Max-Dependency criterion maximizes the mutual information between the discrete data and the classification variable \cite{sharmin2019simultaneous,peng2005feature}. It has been widely used in feature selection \cite{peng2005feature,wan2022r2ci,ren2015learning}, feature weighting \cite{zhang2021attribute} and feature extraction \cite{bishop2006pattern}. Regarding the generalization ability, existing discretizers often choose to maintain a small number of intervals \cite{ramirez2015multivariate,cano2016laim,tahan2018emdid}. However, if the number of intervals is too small, a significant amount of discriminant information will be lost. On the other hand, if it is too large, the resulting discretization scheme may not generalize well to the test data. It is hence difficult to decide the optimal number of intervals. 
In this paper, 
a Min-Divergence criterion is proposed to explicitly maximize the generalization ability by minimizing the divergence between the distribution of training data and that of validation data. This criterion is integrated with the Max-Dependency criterion to form the proposed MDmD criterion, which could achieve a better trade-off between the discriminant power and generalization ability so that the subsequent classifier could work well.

The proposed MDmD criterion is technically appealing but
difficult to be applied in practice, as it is difficult to reliably estimate the high-order joint distributions between attributes and classification variable. To tackle this problem, instead of maximizing the mutual information between all attributes and the classification variable, we propose to maximize the summation of the mutual information between each attribute and the classification variable, also known as Max-Relevance \cite{peng2005feature,sharmin2019simultaneous}. 
At the same time, we propose to minimize the summation of divergences between distributions when one attribute is evaluated at a time. 
These two criteria are combined to form the proposed Max-Relevance-Min-Divergence (MRmD) criterion, which maximizes the discriminant power by maximizing the mutual information between discrete attributes and the classification variable, and simultaneously maximizes the generalization ability by minimizing the divergence between the distributions of training data and validation data. 
{It is time-consuming to exhaustively search for the global optimal solution. Following the design of many discretization methods, \eg, MDLP~\cite{fayyad1993multi}, CAIM~\cite{kurgan2004caim} and Ameva~\cite{gonzalez2009ameva}, a greedy top-down hierarchical splitting algorithm is used together with the proposed MRmD criterion to derive a near-optimal discretization scheme.}

The proposed MRmD criterion 
is integrated with one of the most recent developments of naive Bayes classifier, Regularized Naive Bayes (RNB) \cite{RNB2020shihe}, and compared with the state-of-the-art discretization methods and classifiers on 45 benchmark datasets. The proposed method significantly outperforms the compared methods on most of the datasets.

Our contributions can be summarized as follows.
1) We identify the key limitations of existing discretization methods that they often overemphasize maximizing the discriminant power, which limits the improvement of the generalization ability. 
2) To tackle this problem, a Max-Dependency-Min-Divergence criterion is proposed to simultaneously maximize the discriminant power and minimize the distribution discrepancy so that the derived discretization scheme could generalize well to the data population. 3) To tackle the challenges of reliable estimation of the joint probabilities in MDmD, a more practical solution, Max-Relevance-Min-Divergence discretization scheme, is proposed to derive the optimal discretization scheme for one attribute at a time. 
4) The proposed method is systematically evaluated on {45 benchmark datasets} and demonstrates superior performance compared with the state-of-the-art discretization methods and classifiers.

\section{Related work}
\label{related work}

Discretization methods have been deployed to extract knowledge from data in many machine learning algorithms such as decision tree~\cite{wang2020linear}, rule-based learning~\cite{mu2020parallel} and naive Bayes~\cite{jiang2019class,RNB2020shihe,zhang2021attribute}.
Discretization methods can be categorized according to many properties \cite{ramirez2015multivariate}:

\noindent {\textbf{Local vs. Global:} Local methods \cite{xun2021novel,fayyad1993multi} generate intervals based on partial data, whereas global ones \cite{cano2016laim,ramirez2015multivariate,kurgan2004caim,tsai2008discretization} consider all available data.}

\noindent {\textbf{Dynamic vs. Static:} Dynamic discretizers \cite{sharmin2019simultaneous} interact with learning models whereas static ones \cite{cano2016laim,peker2021application} execute before the learning stage.}

\noindent {\textbf{Splitting vs. Merging:} This relates to the top-down split \cite{xun2021novel,kurgan2004caim,tsai2008discretization} or bottom-up merge \cite{peker2021application} strategy in producing new intervals.}

\noindent {\textbf{Univariate vs. Multivariate:} Univariate algorithms \cite{xun2021novel,cano2016laim,tsai2019optimal} discretize each attribute separately whereas multivariate discretizers \cite{ramirez2015multivariate,tahan2018emdid} consider a combination of attributes when discretizing data.}

\noindent {\textbf{Direct vs. Incremental:} Direct methods \cite{yang2001proportional,ramirez2015multivariate} divide the range into several intervals simultaneously, while incremental ones \cite{cano2016laim,xun2021novel,tsai2008discretization,kurgan2004caim,fayyad1993multi} begin with a simple discretization and improve it gradually using more criteria.}


Depending on whether the class label is used,  discretization methods can be divided into supervised, semi-supervised and unsupervised methods \cite{ramirez2015multivariate}. Equal-width and equal-frequency discretization are representative unsupervised methods~\cite{tsai2019optimal}. Minimal Optimized Description Length is a representative semi-supervised 
method, which applies the Bayesian rule on both labeled and unlabeled data for discretization~\cite{bondu2010non}. Supervised methods can be further divided into wrapper-based methods~\cite{ramirez2015multivariate,tahan2018emdid,tran2017new,zhou2021evolutionary,chen2021feature} and filter-based methods~\cite{xun2021novel,peker2021application,cano2016laim,kurgan2004caim,tsai2008discretization}. The former optimizes the discretization scheme by utilizing the classification feedback~\cite{ramirez2015multivariate,tahan2018emdid,zhou2021evolutionary,tran2017new,chen2021feature}, while the latter optimizes some indirect target for data discretization, \eg, information entropy~\cite{xun2021novel,fayyad1993multi}, mutual information \cite{sharmin2019simultaneous} and interdependency~\cite{cano2016laim,kurgan2004caim,tsai2008discretization}. 

Wrapper-based methods \cite{ramirez2015multivariate,tahan2018emdid,zhou2021evolutionary,tran2017new,chen2021feature} iteratively refine the discretization scheme by using the classification feedback. Evolutionary algorithms are often utilized to discretize data by maximizing the classification accuracy and minimizing the number of intervals \cite{ramirez2015multivariate}. Tahan and Asadi developed an evolutionary multi-objective discretization to handle the imbalanced datasets~\cite{tahan2018emdid}.
Tran \emph{et al.} initialized the discretization scheme by using the MDLP criterion~\cite{fayyad1993multi} and utilized barebones particle swarm optimization to fine-tune the derived scheme  \cite{tran2017new}. In \cite{zhou2021evolutionary}, the particle swarm optimization strategy is used to explore the interaction between features for better discretization. Chen \emph{et al.} developed a genetic algorithm based on the fuzzy rough set to effectively explore the data association \cite{chen2021feature}.

Filter-based methods~\cite{xun2021novel,peker2021application,kurgan2004caim,tsai2008discretization} have been popular in recent years for their strong theoretical background. MDLP is one of the most popular discretization methods in many classifiers \cite{jiang2019class,zhang2021attribute,RNB2020shihe}, which hierarchically partitions data by maximizing the information entropy \cite{fayyad1993multi}. To avoid excessive splitting, it defines a stop criterion derived from channel coding theory. Xun \emph{et al.} developed a multi-scale discretization method to obtain the set of cut points with different granularity and utilized the MDLP criterion to determine the best cut point \cite{xun2021novel}. Other statistical measures have also been widely used in data discretization  \cite{peker2021application,cano2016laim,tsai2019optimal}. Kurgan and Cios developed a CAIM criterion based on a quanta matrix to select boundary points iteratively within a pre-defined number of intervals \cite{kurgan2004caim}. Cano \emph{et al.} extended it for multi-label data \cite{cano2016laim}. Tsai \emph{et al.} introduced a discretization method based on CACC by taking the overall data distribution into account~\cite{tsai2008discretization}. In \cite{rahman2016discretization}, low-frequency values are discretized and the correlation between discrete attribute and continuous attribute is utilized for discretization. Chi-square statistics between the discrete data and the classification variable, \eg, modified Chi2 and extended Chi2, have been recently developed for data discretization \cite{peker2021application}.

Most discretization methods \cite{xun2021novel,cano2016laim,kurgan2004caim,tsai2008discretization,fayyad1993multi} emphasize maximizing the discriminant power, but they pay little attention to the generalization capability, \eg, they often restrict the number of discrete intervals to be small, in the hope of achieving a satisfactory generalization ability. If a discretization method considers maximizing these two simultaneously, 
the subsequent classifier will achieve a better classification performance on novel test data.

\section{Proposed discretization method}
\label{proposed method}
\subsection{Analysis of existing discretization methods}

Data discretization is crucial to improve the learning efficiency and generalization of classifiers \cite{ramirez2015multivariate}. Many methods have been designed to maximize the statistical dependency between discretized features and  classification variables in many different forms, \eg, information gain in MDLP \cite{fayyad1993multi,xun2021novel}, class-attribute dependency in CADD \cite{ching1995class}, and class-attribute interdependency in CAIM \cite{kurgan2004caim,cano2016laim}. But they often neglect the fact that the primary target of data discretization in
classification is to improve generalization ability, \ie, by discretizing similar values into one interval, the data distribution can be better estimated so that it fits well to novel test samples.

MDLP is one of the most widely used discretization methods \cite{jiang2019class,RNB2020shihe,zhang2021attribute}, \eg, it is the default discretization method in Weka toolbox \cite{frank2010weka}.
It hierarchically splits the dynamic range into smaller ones. For each attribute, a cut point $d$ is selected to divide the current set $\mathcal{S}$ into two subsets $\mathcal{S}_1$ and $\mathcal{S}_2$, which maximizes the information gain 
$G(\mathcal{S}, d) = E(\mathcal{S}) -\frac{|\mathcal{S}_1|}{|\mathcal{S}|}E(\mathcal{S}_1) - \frac{|\mathcal{S}_2|}{|\mathcal{S}|}E(\mathcal{S}_2)$, 
where $E(\mathcal{S}) = -\sum\limits_{c \in \mathcal{C}} P(c,\mathcal{S})\log P(c,\mathcal{S})$ is the entropy, $P(c, \mathcal{S})$ is the probability of class $c$ in $\mathcal{S}$ and $\mathcal{C}$ is the set of classes. It can be shown that the accumulative information gain is equivalent to the mutual information between the discrete attribute and the classification variable. Greedily maximizing the information gain may split the attribute into too many small intervals with too few samples so that the likelihood probabilities can not be reliably estimated. To prevent this, MDLP requires $G(\mathcal{S}, d)$ to be greater than a threshold that is derived from the overhead of information encoding, which may not be in line with the classification point of view. It often leads to an early stop during splitting, and hence a significant discriminant information loss.

Class-Attribute Dependent Discretizer (CADD) \cite{ching1995class} maximizes the discriminant information via maximizing $CADD(X, C) = \frac{I(X;C)}{E(X,C)}$, 
where $I(X;C)$ is the mutual information and $E(X,C)$ is the joint entropy. Maximizing CADD tends to produce too many small discretization intervals. To prevent this, a user-specified threshold is utilized to constrain the number of intervals, but with no guarantee of optimality.

The CAIM discretization utilizes a heuristic measure $CAIM(X,C)  = \frac{1}{n}\sum_{i=1}^n\frac{\max_{c\in \mathcal{C}} q_{i,c}^2}{M_i}$ to model the interdependence between classes and attributes \cite{kurgan2004caim,cano2016laim}, 
where $q_{i,c}$ is the number of samples in class $c$ in the $i$-th interval, and $M_i$ is the number of samples in the $i$-th interval. The number of intervals generated by CAIM is often close to the number of classes, which may limit the performance of CAIM, especially when the number of classes is small.

Existing approaches mainly focus on maximizing the discriminant power, which often split attributes into too many small intervals. This defeats the purpose of data discretization in classification, \ie, to improve the generalization. To retain the generalization ability, they often restrict the number of intervals to a predefined number or the number of classes \cite{kurgan2004caim,cano2016laim,tsai2008discretization}, or require the information gain to be larger than a threshold \cite{fayyad1993multi,xun2021novel}. These methods lack a measure to explicitly maximize the generalization ability.

\subsection{Maximal-dependency-minimal-divergence for data discretization}
In this paper, the target is to derive an optimal discretization scheme $\bm{\mathcal{D}}$ to transform continuous attributes $\bm{X}$ into discrete ones $\bm{A}$, which simultaneously maximizes the discriminant power of the discretized data and maximizes the generalization ability to the data not used in training, 
\begin{equation}
\label{eqn:A}
\bm{A} = f_D(\bm{X},\bm{\mathcal{D}}),
\end{equation}
where $f_D$ denotes the discretization function, and $\bm{\mathcal{D}} = \{\mathcal{D}_1, \mathcal{D}_2,...,\mathcal{D}_m\}$ contains the discretization schemes for $m$ features.

\subsubsection{Maximal-dependency criterion}
To maximize the discriminant information, we propose to maximize the mutual information $I(\bm{A};C)$ between the discretized attributes $\bm{A}$ and the classification variable $C$ given the discretization scheme $\bm{\mathcal{D}}$, 
\begin{equation}
    \bm{\mathcal{D}}^* = \argmax_{\bm{\mathcal{D}}} I(\bm{A};C),
\end{equation}
\begin{equation}
I(\bm{A};C) =\sum_{c \in \mathcal{C}} \sum_{\bm{a}\in \bm{A}} P(\bm{a},c)\log\frac{P(\bm{a},c)}{P(\bm{a})P(c)},\label{max_dependency}
\end{equation}
where $P(\bm{a},c)$ is the joint probability distribution and $P(\bm{a})$ and $P(c)$ are the respective marginal probabilities.
This criterion is often known as Max-Dependency \cite{peng2005feature,sharmin2019simultaneous}. 
Apparently, it is difficult to reliably estimate both the joint probability distribution $P(\bm{a},c)$ and the marginal probability distribution $P(\bm{a})$ due to the high dimensionality. Furthermore, the Max-Dependency criterion tends to greedily maximize the discriminant power and hence over-discretize the continuous data into too many small intervals, \ie, each unique value in the numerical attribute may be treated as a separate interval, in which the generalization capability would not be improved.

\subsubsection{Minimal-divergence criterion}
To maximize the generalization ability, we propose to minimize the Jensen-Shannon (JS) divergence~\cite{cover2012elements} between the training data distribution and the test data distribution. As the latter is in general unknown, we hence aim to minimize the distribution discrepancy between training data and validation data instead. The intuition behind is that by minimizing the JS divergence $D_{JS}(P^{t}(\bm{a})||P^{v}(\bm{a}))$ describing the similarity between the distribution $P^{t}(\bm{a})$ of the training data $\bm{A}^{t}$ and the distribution $P^{v}(\bm{a})$ of the validation data $\bm{A}^{v}$, the derived discretization scheme $\bm{\mathcal{D}}$ could generalize well from the training data to the novel test data. Formally, $D_{JS}(P^{t}(\bm{a})||P^{v}(\bm{a}))$ is defined as:
\begin{equation}
D_{JS}(P^{t}(\bm{a})||P^{v}(\bm{a})) = \frac{1}{2}(D_{KL}(P^{t}(\bm{a})||P^{*}(\bm{a})) + D_{KL}(P^{v}(\bm{a})||P^{*}(\bm{a}))),
\label{min_js}
\end{equation}
where $P^*(\bm{a}) = \frac{1}{2}(P^{t}(\bm{a})+P^{v}(\bm{a}))$ 
and $D_{KL}(P^{t}(\bm{a})||P^*(\bm{a}))$ is the Kullback-Leibler divergence between $P^{t}(\bm{a})$ and $P^*(\bm{a})$, 
\begin{equation}
D_{KL}(P^{t}(\bm{a})||P^*(\bm{a}))= \sum_{\bm{a} \in \mathcal{A}} P^t(\bm{a})\log \frac{P^t(\bm{a})}{P^*(\bm{a})}.
\end{equation}
Similarly, $D_{KL}(P^{v}(\bm{a})||P^*(\bm{a}))$ is defined as:
\begin{equation}
D_{KL}(P^{v}(\bm{a})||P^*(\bm{a}))= \sum_{\bm{a} \in \mathcal{A}} P^v(\bm{a})\log \frac{P^v(\bm{a})}{P^*(\bm{a})}.
\end{equation}
$P^{t}(\bm{a})$, $P^{v}(\bm{a})$ and $P^*(\bm{a})$ are the probability distributions of $\bm{a}$ given the attribute set $\bm{A}^{t}$, $\bm{A}^{v}$ and $\bm{A}^*$ respectively. 
The JS divergence has been utilized as a distance metric between two distributions. It is symmetric, \ie, $D_{JS}(P^{t}(\bm{a})||P^{v}(\bm{a})) = D_{JS}(P^{v}(\bm{a})||P^{t}(\bm{a}))$. $D_{JS}(P^{t}(\bm{a})||P^{v}(\bm{a})) \in [0,1]$. 
The smaller JS divergence represents the higher similarity between these two distributions, and hence the derived discretization scheme could generalize well to the novel test data. The Minimal-Divergence criterion is hence defined as:
\begin{equation}
    \bm{\mathcal{D}}^* = \argmin_{\bm{\mathcal{D}}} D_{JS}(P^{t}(\bm{a})||P^{v}(\bm{a})).
\end{equation}

\subsubsection{Maximal-dependency-minimal-divergence criterion}
To simultaneously maximize the discriminant power and the generalization ability of the discretized attributes, we propose to maximize the dependency $I(\bm{A};C)$ between discrete attributes $\bm{A}$ and classification variable $C$, and minimize the divergence $D_{JS}(P^{t}(\bm{a})||P^{v}(\bm{a}))$ between the distribution of training data and that of validation data given the discretization scheme $\bm{\mathcal{D}}$,
\begin{equation}
\label{eqn:MDmD}
\bm{\mathcal{D}}^* = \argmax_{\bm{\mathcal{D}}} \lambda I(\bm{A};C) -  D_{JS}(P^{t}(\bm{a})||P^{v}(\bm{a})),
\end{equation}
where $\lambda$ is the hyper-parameter to balance the two terms. Note that the two terms compete with each other. On the one hand, when all the data are discretized into one bin, $I(\bm{A};C) = 0$, indicating that the discriminant information is totally lost, but $D_{JS}(P^{t}(\bm{a})||P^{v}(\bm{a})) = 0$, \ie, the two distributions are identical, and hence the generalization is maximized. On the other hand, when each unique sample is discretized into a separate bin, the discriminant information is maximized, while it {does not} improve the generalization ability. The proposed MDmD criterion provides a solution to find the optimal trade-off between the discriminant power and the generalization ability. 

\subsection{Maximal-relevance-minimal-divergence criterion for data discretization}
The proposed MDmD criterion is technically appealing but difficult to implement in practice, as it is hard to reliably estimate the high-order joint distribution $P(\bm{a},c)$, $P^t(\bm{a})$ and $P^{v}(\bm{a})$. 
{Inspired by~\cite{peng2005feature}, we propose the Max-Relevance criterion for data discretization. More specifically, following the chain rule of mutual information~\cite{cover2012elements}, $I(\bm{A};C)=\sum_{j=1}^m I(A_j;C|A_{j-1},\cdots,A_1)$, where $I(A_j;C|A_{j-1},\cdots,A_1)$ is the conditional mutual information between $A_j$ and $C$ conditioned on $A_{j-1},\cdots,A_1$. If we ignore the high-order interaction between features, \ie, $I(A_j;C|A_{j-1},\cdots,A_1)\approx I(A_j;C)$, we have $I(\bm{A};C)\approx\sum_{j=1}^m I(A_j;C)$,  where $I(A_j;C)$ is the mutual information between $A_j$ and $C$. The detailed derivation of this approximation can be found in~\cite{cover2012elements}. The Max-Relevance criterion is then given as follows,
\begin{equation}
\label{eqn:MR}
\bm{\mathcal{D}}^* = \argmax_{\bm{\mathcal{D}}}  \sum_{j=1}^m I({A}_j;C),
\end{equation}
which has been widely used to approximate the Max-Dependency criterion \cite{peng2005feature,sharmin2019simultaneous,wan2022r2ci}.}

{For the second term in Eqn. (\ref{eqn:MDmD}), instead of estimating the divergence between $P^{t}(\bm{a})$ and $P^{v}(\bm{a})$ jointly considering all the attributes, $D_{JS}(P^{t}(\bm{a})||P^{v}(\bm{a}))$ can be estimated  by considering them one by one. Following the chain rule of divergence~\cite{cover2012elements}, $D_{KL}(P^{t}(\bm{a})||P^{*}(\bm{a})) = \sum_{j=1}^m D_{KL}(P^t(a_j|a_{j-1},\cdots,a_1)||P^*(a_j|a_{j-1},\cdots,a_1))$, where $D_{KL}(P^t(a_j|a_{j-1},\cdots,a_1)||P^*(a_j|a_{j-1},\cdots,a_1))$ is the conditional divergence. If we ignore the high-order interaction between features, \ie, $P^t(a_j|a_{j-1},\cdots,a_1) \approx P^t(a_j)$ and $P^*(a_j|a_{j-1},\cdots,a_1) \approx P^*(a_j)$, we have $D_{KL}(P^{t}(\bm{a})||P^{*}(\bm{a})) \approx \sum_{j=1}^m D_{KL}(P^t(a_j)||P^*(a_j))$. It is then easy to show that $D_{JS}(P^{t}(\bm{a})||P^{v}(\bm{a})) \approx \sum_{j=1}^m D_{JS}(P^t(a_j)||P^v(a_j))$, where $D_{JS}(P^t(a_j)||P^v(a_j))$ is the JS divergence between  training data distribution and validation data distribution for the $j$-th attribute given the discretization scheme $\mathcal{D}_j$. The Min-Divergence criterion can hence be simplified as,}
\begin{equation}
\label{eqn:mD}
\bm{\mathcal{D}}^* = \argmin_{\bm{\mathcal{D}}}  \sum_{j=1}^m D_{JS}(P^{t}(a_j)||P^{v}(a_j)),
\end{equation}

We combine Eqn. (\ref{eqn:MR}) and Eqn. (\ref{eqn:mD}) to form the proposed Maximal-Relevance-Minimal-Divergence (MRmD) criterion,
\begin{equation}
\label{eqn:MRmD}
\bm{\mathcal{D}}^* = \argmax_{\bm{\mathcal{D}}}  \sum_{j=1}^m \left[ \lambda I({A}_j;C) -  D_{JS}(P^{t}(a_j)||P^{v}(a_j))\right]. 
\end{equation}
{Given the discretization scheme $\bm{\mathcal{D}}$, each original feature $\bm{x}_j$ can be discretized into $A_j$ as in Eqn.~\eqref{eqn:A}, and then the mutual information $I(A_j;C)$ and JS divergence $D_{JS}(P^t(a_j)||P^v(a_j))$ can be estimated.} 
The proposed MRmD aims to find the optimal scheme $\bm{\mathcal{D}}^*$ that maximizes the relevance of discretized attributes with respect to the classification variable via the first term, and maximizes the generalization ability by minimizing the distribution discrepancy between training data and validation data via the second term.

To derive $\bm{\mathcal{D}}^*$, 
it is not difficult to show that each attribute $A_j$ can be processed separately to derive its optimal discretization scheme $\mathcal{D}_j^*$,
\begin{equation}
\label{eqn:DJStar}
{\mathcal{D}}_j^* = \argmax_{{\mathcal{D}}_j} {\Psi}({A}_j;C),
\end{equation}
\begin{equation}
{\Psi}({A}_j;C)=  \lambda I({A}_j;C) - D_{JS}(P^{t}(a_j)||P^{v}(a_j)).
   \label{eqn:psi}
\end{equation}
After deriving the optimal solution for each attribute, the optimal discretization scheme is obtained as 
$\bm{\mathcal{D}}^* = \{\mathcal{D}_1^*, \mathcal{D}_2^*,...,\mathcal{D}_m^*\}$. 

\subsection{Proposed MRmD data discretization}

\begin{figure}[!t]
	\centering
	\includegraphics[width=1\textwidth]{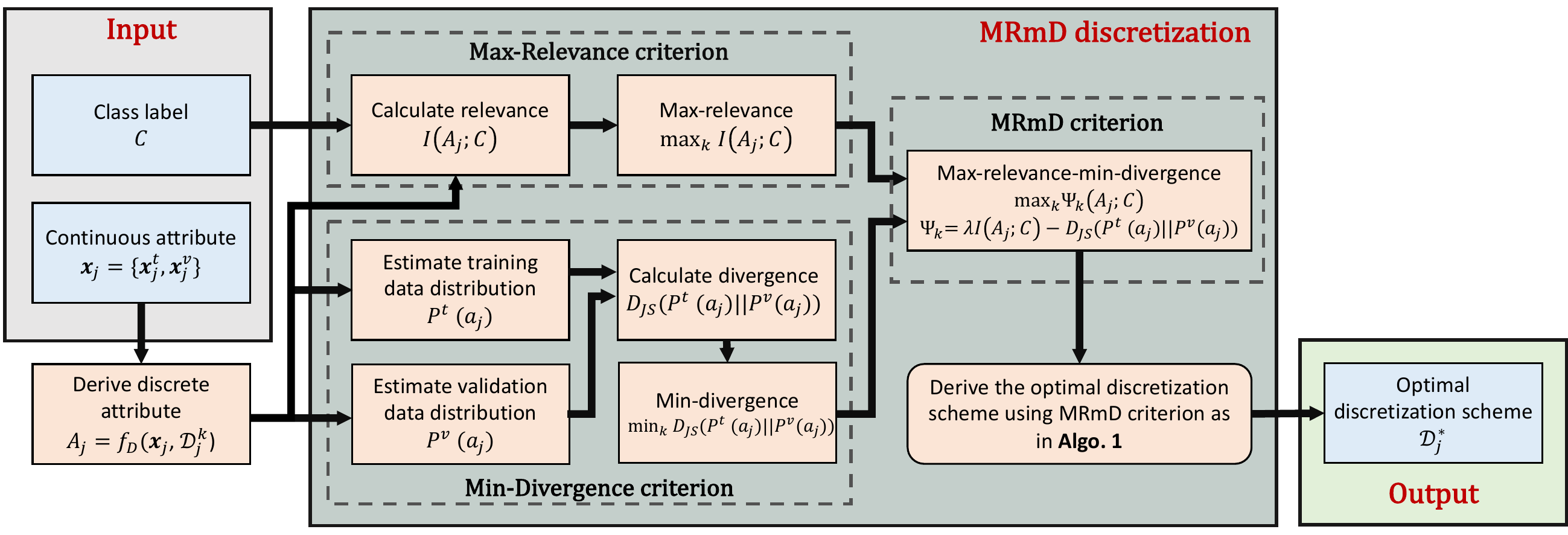}
	\caption{The proposed MRmD discretization framework. The optimal set of cut points for each attribute is selected by maximizing the proposed MRmD criterion, where the Max-Relevance criterion maximizes the discriminant information of the discretized data, and the Min-Divergence criterion maximizes the generalization by minimizing the distribution discrepancy between training and validation data. More details are given in Algo. \ref{alg}.}
	\label{frame}
\end{figure}

An MRmD discretization method is proposed to discretize {the attributes one at a time}, 
with the block diagram shown in Fig.~\ref{frame}. 
For each continuous attribute, the proposed method iteratively determines the cut points for discretization by maximizing the MRmD criterion, where the Max-Relevance criterion is achieved by maximizing the mutual information between the discretized attribute and the classification variable, and the Min-Divergence criterion is achieved by minimizing the distribution discrepancy between training data and validation data. The two criteria are combined as the MRmD criterion, to derive an optimal set of cut points to discretize the attributes. 


Following the design in MDLP \cite{xun2021novel,fayyad1993multi} and many others \cite{kurgan2004caim,tsai2008discretization,cano2016laim}, a greedy top-down hierarchical splitting paradigm is designed to derive the optimal solution. 
More specifically, for each attribute $\bm{x}_j$, we initialize its discretization scheme $\mathcal{D}_j^*$ as an empty set, 
and treat the whole dynamic range initially as one interval. Hence $I({A}_j;C) = 0$ and $D_{JS}(P^{t}(a_j)||P^{v}(a_j)) = 0$. 
{The optimal cut points $\mathcal{D}_j^*$ are selected from a candidate set $\mathcal{S}_j$, which is initialized as 
$\mathcal{U}(\bm{x}_j)$, the unique values of $\bm{x}_j$.} 
It can be shown that the number of possible discretization schemes for $\bm{x}_j$ is $2^{|\mathcal{S}_j|}$. 
It is expensive to exhaustively evaluate every feasible discretization scheme. 
$\forall d_k \in \mathcal{S}_j$, $\mathcal{D}_j^k = \mathcal{D}_j^* \cup d_k$, {and we use $\mathcal{D}_j^k$ to discretize $\bm{x}_j$, and evaluate the MRmD criterion ${\Psi}_k$ defined in Eqn. \eqref{eqn:psi} for every $d_k$. Then, we select the cut point $d_{k_{max}}$ that maximizes ${\Psi}_k$, and update the optimal discretization scheme as $\mathcal{D}_j^* = \mathcal{D}_j^* \cup d_{k_{max}}$. The candidate set is updated as $\mathcal{S}_j = \mathcal{S}_j - d_{k_{max}}$.} The proposed method incrementally selects the cut point to divide the dynamic range into intervals until the criterion defined in Eqn. \eqref{eqn:psi} does not increase anymore. 

The proposed MRmD discretization is summarized in Algo.~\ref{alg}. the proposed MRmD discretizes the continuous attribute $\bm{x}_j \in \bm{X}$ one by one. These discretization schemes $\mathcal{D}_j^*$ for all attributes form the complete discretization scheme $\bm{\mathcal{D}}^*$. The proposed MRmD generates a discretization scheme that simultaneously maximizes the discriminant information and the generalization ability, and hence improves the classification performance.

\begin{algorithm}[!ht]
	\scriptsize
	\renewcommand{\algorithmicrequire}{\textbf{Input:} }
	\renewcommand{\algorithmicensure}{\textbf{Output:}}
	\caption{The proposed MRmD discretization scheme.} 
	\begin{algorithmic}[1]
		\Require Input data $\bm{X} = \{\bm{X}^{t}, \bm{X}^{v}\}$ with class label $\bm{c}$ for $\bm{X}$
		\Ensure Discretization scheme $\bm{\mathcal{D}}^* =  \{\mathcal{D}_1^*, \mathcal{D}_2^*,..., \mathcal{D}_m^*\}$
		\State $\bm{\mathcal{D}}^* \gets \emptyset$ \Comment{Initialize $\bm{\mathcal{D}}^*$ as an empty set}
		\For{$\bm{x}_j \in \bm{X}$} \Comment{Loop through all attributes}
		\State $\mathcal{D}_j^* \gets \emptyset$ \Comment{Initialize $\mathcal{D}_j^*$ as an empty set}
		\State $\mathcal{S}_j \gets \mathcal{U}(\bm{x}_j)$ \Comment{Initialize $\mathcal{S}_j$ as the set of unique values}
 		\State ${\Psi}_{{max}} \gets -\infty$ \Comment{Initialize optimal ${\rm MRmD}$ value}
		\While{$({\mathcal{S}}_j \ne {\emptyset})$} 
		\Comment{Incrementally select the cut point}
		\For {$d_k \in {\mathcal{S}}_j$} \Comment{For each possible cut point}
		\State $\mathcal{D}_j^k = \mathcal{D}_j^* \cup d_k$ \Comment{Include $d_k$ as the cut point}
		\State $A_j = f_D(\bm{x}_j,{D}_j^k)$ \Comment{Discretize $\bm{x}_j$ using $\mathcal{D}_j^k$}
		\State ${\Psi}_k = \lambda I({A}_j;C) - D_{JS}(P^{t}(a_j)||P^{v}(a_j))$
		\Comment{Calculate ${\Psi}_k$}
		\EndFor
        \State $k_{max} = \argmax_k {\Psi}_{k}$ \Comment{Derive $d_{k_{max}}$ with maximal ${\Psi}_{k_{max}}$}
        \If{${\Psi}_{k_{max}} \leq {\Psi}_{max}$}
            \State break;
        \EndIf
		\State ${\Psi}_{max} \gets {\Psi}_{k_{max}}$ \Comment{Update ${\Psi}_{max}$}
		\State $\mathcal{D}_j^* \gets \mathcal{D}_j^* \cup d_{k_{max}}$ \Comment{Update $\mathcal{D}_j^*$}
		\State $\mathcal{S}_j \gets \mathcal{S}_j - d_{k_{max}}$ \Comment{Update $\mathcal{S}_j$}
        \EndWhile
		\State $\bm{\mathcal{D}}^* \gets \bm{\mathcal{D}}^* \cup \mathcal{D}_j^*$ \Comment{Add $\mathcal{D}_j^*$ into $\bm{\mathcal{D}}^*$}
		\EndFor
		\State \Return $\bm{\mathcal{D}}^*$
	\end{algorithmic}
	
	\label{alg}
\end{algorithm}

\subsection{Analysis of hyper-parameter $\lambda$}
The hyper-parameter $\lambda$ plays an important role in the proposed MRmD. As discussed early, the MRmD value ${\Psi}({A}_j;C)$ is initialized as zero at the beginning of the top-down discretization. Both terms in Eqn. (\ref{eqn:psi}) increases with the number of discretization intervals. It is hence difficult to derive the optimal MRmD value. 
To tackle this problem, $\lambda$ is defined as,
\begin{equation}
\label{eqn:lambda}
\lambda = e^{-\frac{|\mathcal{D}_j^*|}{N_D}},
\end{equation}
where $|\mathcal{D}_j^*|$ is the number of cut points in the current discretization scheme, and $N_D$ is empirically set to 50. {The designed weighting function satisfies the following properties: 1). The value of $\lambda$ is between 0 and 1. 2) $\lambda$ monotonically decreases with the number of cut points.} 
In the earlier stage, when there are only a small number of cut points in the discretization scheme, $\lambda$ is large and hence more emphasis is put on the first term in Eqn. \eqref{eqn:psi}, to highlight the importance of maximizing the discriminant information. As more cut points are added into the discretization scheme, $\lambda$ becomes smaller and then more emphasis is put on the second term of Eqn. \eqref{eqn:psi} so that more emphasis is put on 
improving the generalization performance. Such a design could help the proposed MRmD discretization generate a discretization scheme that achieves an optimal trade-off between the generalization capability and the discrimination information for the discretized data.



\section{Experimental results}
\label{exp}
\subsection{Experimental settings}
The proposed MRmD discretization is compared with 
five popular filter-based discretization methods, Ameva~\cite{gonzalez2009ameva}, CAIM \cite{kurgan2004caim}, MDLP~\cite{fayyad1993multi}, Modified Chi2~\cite{tay2002modified} and PKID~\cite{yang2001proportional}, and a recent wrapper-based approach, EMD~\cite{ramirez2015multivariate}. 
Ameva~\cite{gonzalez2009ameva} and CAIM \cite{kurgan2004caim} are two popular methods for discretization in credit scoring models for operational research~\cite{silva2022class}. MDLP \cite{fayyad1993multi} and PKID \cite{yang2001proportional} are widely used in NB classifiers~\cite{RNB2020shihe,jiang2019class,zhang2021attribute} and feature selection~\cite{tran2017new}. Modified Chi2 \cite{tay2002modified} has been recently used to improve the ensemble classification methods~\cite{peker2021application}. EMD \cite{ramirez2015multivariate} has been used in many applications recently, \eg, high-resolution remote sensing~\cite{chen2021feature} and feature selection~\cite{zhou2021evolutionary}. {Two classifiers are used for evaluation, naive Bayes classifier and decision tree (C4.5)~\cite{quinlan2014c4}.}


The proposed MRmD is then integrated with one of the recent naive Bayes classifiers, RNB (regularized naive Bayes) \cite{RNB2020shihe}, denoted as MRmD-RNB. It is compared with state-of-the-art NB classifiers including RNB~\cite{RNB2020shihe}, WANBIA~\cite{zaidi2013alleviating}, CAWNB~\cite{jiang2019class} and AIWNB~\cite{zhang2021attribute}.
{It is also compared with three deep-learning models, ResNet \cite{he2016deep}, FTT \cite{gorishniy2021revisiting} and PWedRVFL \cite{shi2022weighting}. ResNet and WPedRVFL are implemented using the codes provided by the authors of \cite{shi2022weighting}, and FTT is implemented following \cite{gorishniy2021revisiting}.}


\begin{table}[!h]
\caption{Description of compared methods: six discretization methods, four state-of-the-art naive Bayes classifiers and three deep-learning models.}
\scriptsize
\centering
\begin{tabular}{@{}|p{2cm}<\centering|p{10.5cm}|@{}}
\hline
\multicolumn{2}{|c|}{\textbf{Discretization methods}} \\ \hline
Ameva \cite{gonzalez2009ameva,silva2022class}                & Filter-based statistical top-down discretization, maximizing the contingency coefficient based on Chi-square statistics.  \\ \hline
CAIM \cite{kurgan2004caim,cano2016laim}                 & Filter-based statistical top-down discretization, heuristically maximizing the class-attribute interdependency by using quanta matrix. \\ \hline
MDLP~\cite{xun2021novel,fayyad1993multi,RNB2020shihe}                & Filter-based entropy-based top-down discretization, maximizing the information gain using the minimum description length principle.               \\ \hline
Modified Chi2~\cite{peker2021application,tay2002modified}       & Filter-based statistical bottom-up discretization, merging intervals 
dynamically by using the rough set theory.                      \\ \hline
PKID \cite{yang2001proportional}                & Filter-based unsupervised discretization, adjusting the number and size of intervals proportional to the number of training instances. 
\\ \hline
EMD \cite{ramirez2015multivariate,tahan2018emdid} & Wrapper-based multivariate discretization, minimizing the classification error and the number of intervals using genetic algorithm.  \\ \hline
    \multicolumn{2}{|c|}{\textbf{Naive Bayes classifiers}} \\ \hline
RNB~\cite{RNB2020shihe}   & Wrapper-based attribute-weighting naive Bayes, simultaneously optimizing class-dependent and class-independent weights by using the L-BFGS algorithm.   \\ \hline
WANBIA~\cite{zaidi2013alleviating} & Wrapper-based attribute-weighting naive Bayes, optimizing class-independent attribute weights by using the L-BFGS algorithm.            \\ \hline
CAWNB~\cite{jiang2019class}&Wrapper-based attribute-weighting naive Bayes, optimizing class-specific attribute weights by using the L-BFGS algorithm. \\ \hline
AIWNB~\cite{zhang2021attribute}& Filter-based attribute and instance-weighting naive Bayes, combining correlation-based attribute weights with frequency-based instance weights using eager learning AIWNB$^E$ and similarity-based instance weights using lazy learning AIWNB$^L$.\\ \hline
\multicolumn{2}{|c|}{{\textbf{Deep-learning models}}} \\ \hline
{ResNet \cite{he2016deep}}& {Adapted residual networks using 2 or 3 residual blocks.}\\ \hline
{FTT \cite{gorishniy2021revisiting}}& {Adapted transformer with feature tokenizer.} \\ \hline
{PWedRVFL~\cite{shi2022weighting}}& {Combination of pruning-based and weighting-based ensemble deep random vector functional link neural network with re-normalization.} \\ \hline
\end{tabular}
\label{para}
\end{table}

The experiments are conducted on {45 benchmark datasets} in various fields including healthcare, biology, disease diagnosis and business. The datasets are extracted from the UCI machine learning repository{\footnote{https://archive.ics.uci.edu/ml/index.php}}, which have been widely used to evaluate discretization algorithms \cite{ramirez2015multivariate,sharmin2019simultaneous,tsai2019optimal} and naive Bayes classifiers \cite{RNB2020shihe,jiang2019class,zhang2021attribute}. Most datasets are collected from real-world problems. 
The number of instances is distributed between 106 and 10992 and the number of attributes is distributed between 2 and 90. Some datasets contain missing values which are replaced by the mean or mode of the corresponding attribute. Besides, there are both nominal attributes and numerical attributes in some datasets. These 45 datasets provide a comprehensive evaluation of the proposed methods. The statistics of these datasets are summarized in Table~\ref{dataset}. Similarly as in~\cite{ramirez2015multivariate,RNB2020shihe,zhang2021attribute,jiang2019class}, the classification accuracy of each method on each dataset is derived via stratified 10-fold cross-validation. For the proposed method, only 8 out of 9 folds of training data are used in training while the remaining one fold serves as validation data.
\begin{table}[!t]
\caption{Statistics of the benchmark datasets, where Inst., Attr., Class, Num., Nom. and Missing denote the number of instances, attributes, classes, numerical attributes, nominal attributes and whether the dataset contains missing values, respectively. The number of instances is distributed between 106 and 10992, which provides a comprehensive evaluation on datasets of different sizes. The number of attributes also varies across datasets. There are both numerical and nominal attributes, and/or missing values in some of the datasets, which makes the design of discretization scheme and subsequent classifier more challenging.}
\centering
	\resizebox{1.\linewidth}{!}{
\begin{tabular}{@{}ccccccc|ccccccc@{}}
\toprule
              & Inst. & Attr. & Class & Num. & Nom. & Missing 	&     \qquad         & Inst. & Attr. & Class & Num. & Nom. & Missing \\  \midrule
abalone       & 4174  & 8     & 28    & 7    & 1    & N       	& movement      & 360   & 90    & 15    & 90   & 0    & N       \\
appendicitis  & 106   & 7     & 2     & 7    & 0    & N       	& newthyroid    & 215   & 5     & 3     & 5    & 0    & N       \\
australian    & 690   & 14    & 2     & 8    & 6    & N       	& pageblocks    & 5472  & 10    & 5     & 10   & 0    & N       \\
auto          & 205   & 25    & 6     & 15   & 10   & Y       	& penbased      & 10992 & 16    & 10    & 16   & 0    & N       \\
balance       & 625   & 4     & 3     & 4    & 0    & N       	& phoneme       & 5404  & 5     & 2     & 5    & 0    & N       \\
banana        & 5300  & 2     & 2     & 2    & 0    & N       	& pima          & 768   & 8     & 2     & 8    & 0    & N       \\
bands         & 539   & 19    & 2     & 19   & 0    & Y       	& saheart       & 462   & 9     & 2     & 8    & 1    & N       \\
banknote      & 1372  & 5     & 2     & 5    & 0    & N       	& satimage      & 6435  & 36    & 7     & 36   & 0    & N       \\
bupa          & 345   & 6     & 2     & 6    & 0    & N       	& segment       & 2310  & 19    & 7     & 19   & 0    & N       \\
clevland      & 303   & 13    & 5     & 13   & 0    & Y       	& seismic       & 2584  & 19    & 2     & 15   & 4    & N       \\
climate       & 540   & 18    & 2     & 18   & 0    & N       	& sonar         & 208   & 60    & 2     & 60   & 0    & N       \\
contraceptive & 1473  & 9     & 3     & 9    & 0    & N       	& spambase      & 4597  & 57    & 2     & 57   & 0    & N       \\
crx           & 690   & 15    & 2     & 6    & 9    & Y       	& specfheart    & 267   & 44    & 2     & 44   & 0    & N       \\
dermatology   & 366   & 34    & 6     & 34   & 0    & Y       	& tae           & 151   & 5     & 3     & 5    & 0    & N       \\
ecoli         & 336   & 7     & 8     & 7    & 0    & N       	& thoracic      & 470   & 17    & 2     & 3    & 14   & N       \\
flare-solar   & 1066  & 9     & 2     & 9    & 0    & N       	& titanic       & 2201  & 3     & 2     & 3    & 0    & N       \\
glass         & 214   & 9     & 7     & 9    & 0    & N       	& transfusion   & 748   & 5     & 2     & 5    & 0    & N       \\
haberman      & 306   & 3     & 2     & 3    & 0    & N       	& vehicle       & 846   & 18    & 4     & 18   & 0    & N       \\
hayes         & 160   & 4     & 3     & 4    & 0    & N       	& vowel         & 990   & 13    & 11    & 13   & 0    & N       \\
heart         & 270   & 13    & 2     & 13   & 0    & N       	& wine          & 178   & 13    & 3     & 13   & 0    & N       \\
hepatitis     & 155   & 19    & 2     & 19   & 0    & Y       	& wisconsin     & 699   & 9     & 2     & 9    & 0    & N       \\
iris          & 150   & 4     & 3     & 4    & 0    & N       	& yeast         & 1484  & 8     & 10    & 8    & 0    & N       \\ 
mammographic  & 961   & 5     & 2     & 5    & 0    & N       	& &&&&&& \\ \bottomrule
\end{tabular}
}
\label{dataset}
\end{table}

\subsection{Comparisons to state-of-the-art discretization methods}
The proposed discretization is compared with state-of-the-art discretization methods, Ameva~\cite{gonzalez2009ameva}, CAIM~\cite{kurgan2004caim}, MDLP~\cite{fayyad1993multi}, Modified Chi2~\cite{tay2002modified}, PKID~\cite{yang2001proportional} and EMD~\cite{ramirez2015multivariate} on the 45 datasets. 
The results are summarized in Table~\ref{exp-nb}, where the results of the compared methods are obtained by using the KEEL tool~\cite{alcala2009keel}. 
The highest classification accuracy on each dataset among all compared methods is highlighted in bold. 
The average classification accuracy over all datasets is summarized at the bottom of Table~\ref{exp-nb}.  
\begin{table}[!ht]
\tiny
\caption{\footnotesize{{Comparisons of different discretization methods under the naive Bayes classification framework. The proposed MRmD achieves the best average classification accuracy, and outperforms the second best method, EMD \cite{ramirez2015multivariate}, by 1.69\% on average.}}}
\centering
\resizebox{0.9\columnwidth}{!}{
\begin{tabular}{cccccccc}
              \toprule
              & \multicolumn{1}{c}{Ameva \cite{gonzalez2009ameva}} & \multicolumn{1}{c}{CAIM \cite{kurgan2004caim}} & \multicolumn{1}{c}{MDLP \cite{fayyad1993multi}} & \multicolumn{1}{c}{Modified Chi2~\cite{tay2002modified}} & \multicolumn{1}{c}{PKID \cite{yang2001proportional}} & \multicolumn{1}{c}{EMD \cite{ramirez2015multivariate}} & \multicolumn{1}{c}{MRmD}           \\ \midrule
abalone       & 21.27±3.12          & 25.85±1.52          & 24.96±1.72          & 24.29±1.88          & \textBF{26.11±2.25} & 25.78±2.39           & 25.54±1.70           \\
appendicitis  & \textBF{88.00±9.58} & 87.09±10.61         & 87.09±9.70          & 85.18±10.71         & 86.09±9.89          & 87.09±10.06          & 87.91±12.10          \\
australian    & 85.07±3.99          & \textBF{86.38±4.64} & 84.49±4.32          & 84.49±4.16          & 85.51±3.68          & 85.65±3.46           & 86.37±3.86           \\
autos         & 67.29±11.46         & 64.97±8.73          & 67.32±11.80         & 64.88±11.86         & 72.69±11.12         & 66.06±6.71           & \textBF{76.93±10.75} \\
balance       & 79.68±4.07          & 80.31±4.23          & 72.66±6.53          & 90.88±1.50          & \textBF{91.20±1.33} & 85.44±3.81           & 91.04±1.70           \\
banana        & 70.49±2.34          & 60.49±2.72          & 72.47±2.22          & 63.00±1.99          & 71.47±2.06          & \textBF{73.57±1.99}  & 72.96±2.44           \\
bands         & 72.55±4.31          & 66.43±3.94          & 50.45±9.07          & 72.35±6.18          & 68.65±7.04          & 65.50±5.52           & \textBF{73.64±6.42}  \\
banknote      & 89.43±2.24          & 89.07±2.31          & 92.05±1.55          & 91.40±2.33          & 92.20±2.27          & \textBF{94.24±1.43}  & 91.55±1.64           \\
bupa          & 65.99±12.29         & 61.69±8.91          & 57.15±7.40          & 63.76±4.60          & 62.77±9.86          & 65.48±8.94           & \textBF{68.42±7.29}  \\
cleveland     & 57.78±4.53          & 56.12±7.58          & 55.45±4.07          & 54.82±7.24          & 55.44±7.72          & 57.09±7.17           & \textBF{58.07±8.89}  \\
climate       & 91.11±1.61          & 91.67±1.71          & 93.52±2.52          & 91.30±2.87          & 90.19±2.49          & \textBF{93.52±2.38}  & 93.51±3.38           \\
contraceptive & 50.64±4.73          & 49.63±2.89          & 50.51±4.84          & 50.24±3.15          & 50.92±2.83          & 52.00±3.44           & \textBF{52.21±3.02}  \\
crx           & 85.51±4.78          & 86.09±4.28          & 85.65±4.80          & 84.20±3.09          & 85.22±3.60          & 85.36±5.89           & \textBF{86.23±6.25}  \\
dermatology   & 98.10±2.23          & 97.55±2.99          & 97.82±2.14          & \textBF{98.65±1.91} & 97.82±2.51          & 94.82±3.30           & 98.63±1.95           \\
ecoli         & 81.27±7.06          & 80.94±4.68          & 82.16±6.16          & 79.48±6.42          & 80.38±6.02          & 78.89±5.63           & \textBF{84.28±5.05}  \\
flare         & 65.57±4.94          & 65.57±4.94          & 67.54±3.80          & 65.29±4.94          & 65.48±4.93          & 67.26±4.09           & \textBF{68.29±5.43}  \\
glass         & 46.47±6.61          & 70.34±14.11         & 72.06±8.38          & 71.99±8.24          & 72.46±9.83          & 71.24±12.34          & \textBF{75.67±9.03}  \\
haberman      & 74.78±6.74          & 73.52±4.56          & 72.85±3.70          & 72.20±4.51          & 72.82±5.45          & 74.49±5.85           & \textBF{74.80±4.29}  \\
hayes         & 74.37±9.73          & 74.37±9.73          & 52.02±8.12          & 79.37±14.29         & 79.32±12.68         & \textBF{81.57±11.15} & 80.09±11.34          \\
heart         & 83.70±8.59          & 84.07±7.21          & 84.07±8.91          & 82.96±9.43          & \textBF{84.44±7.16} & 82.96±8.31           & 84.07±8.91           \\
hepatitis     & 81.96±9.64          & 83.88±10.37         & 83.83±11.62         & 82.63±12.40         & 80.71±11.69         & 82.54±10.21          & \textBF{86.54±11.19} \\
iris          & 93.33±4.44          & 94.00±4.92          & 92.67±3.78          & 93.33±4.44          & 92.00±4.22          & \textBF{95.33±4.27}  & 94.00±7.34           \\
mammographic  & 81.48±4.38          & 82.62±4.04          & 82.21±4.46          & 82.63±5.19          & 83.25±5.48          & \textBF{83.57±5.34}  & 83.36±4.16           \\
movement      & 65.00±6.57          & 65.28±5.75          & 60.56±5.68          & 62.50±7.99          & 66.67±4.90          & 55.83±8.91           & \textBF{68.90±8.84}  \\
newthyroid    & 95.35±3.18          & 95.82±4.16          & 94.89±4.13          & 95.82±4.68          & 96.75±3.17          & 94.94±4.33           & \textBF{97.66±3.96}  \\
pageblocks    & 94.01±0.77          & 93.42±0.65          & 93.11±0.94          & 93.75±0.83          & 91.58±0.96          & 94.06±0.87           & \textBF{94.10±0.85}  \\
penbased      & 86.08±0.91          & 87.12±0.77          & 87.66±0.97          & 87.71±0.87          & 87.22±0.84          & 87.08±0.88           & \textBF{88.67±0.86}  \\
phoneme       & 78.89±1.95          & 78.94±1.82          & 76.89±2.14          & 77.05±1.33          & 77.42±2.18          & \textBF{79.35±2.32}  & 79.13±2.30           \\
pima          & 72.80±4.34          & 73.20±6.04          & 75.26±3.77          & 73.97±4.70          & 74.10±5.04          & \textBF{77.22±3.40}  & 74.61±3.58           \\
saheart       & 65.82±5.54          & 70.35±4.80          & 66.24±5.78          & 67.77±7.94          & 67.56±5.78          & 70.79±3.25           & \textBF{70.79±3.58}  \\
satimage      & 25.28±0.65          & 81.69±1.60          & 82.10±1.31          & 82.22±1.48          & 82.11±1.42          & 81.99±1.56           & \textBF{82.28±1.37}  \\
segment       & 91.26±1.08          & 90.39±1.19          & 91.04±1.59          & 89.87±2.20          & 89.09±2.74          & \textBF{93.55±1.26}  & 92.29±1.69           \\
seismic       & 82.24±2.76          & 81.96±2.87          & 82.00±2.24          & 85.80±1.62          & 82.47±1.68          & 93.34±0.24           & \textBF{93.42±0.01}  \\
sonar         & 77.88±9.10          & 77.45±8.30          & 76.88±12.68         & 78.36±8.25          & 74.52±14.03         & 73.93±10.33          & \textBF{78.40±10.08} \\
spambase      & 89.95±1.60          & 89.38±1.18          & 89.89±1.40          & 90.21±1.43          & 89.45±1.59          & \textBF{92.28±1.52}  & 90.53±1.75           \\
specfheart    & 76.44±8.65          & 76.82±9.62          & 73.05±8.95          & 74.96±9.17          & 77.52±8.40          & \textBF{81.28±3.72}  & 79.71±6.80           \\
tae           & 51.13±15.38         & 49.04±17.35         & 34.42±2.36          & 55.71±10.84         & 49.04±17.63         & 54.38±10.92          & \textBF{56.57±15.45} \\
thoracic      & 81.91±3.95          & 82.13±2.72          & 82.13±3.04          & 82.98±3.81          & 80.43±5.28          & 82.13±3.04           & \textBF{82.98±3.47}  \\
titanic       & 78.10±3.02          & 77.83±2.97          & 77.60±3.22          & 77.88±3.02          & 77.88±3.02          & \textBF{78.33±3.07}           & 78.19±2.34  \\
transfusion   & 76.87±6.65          & 76.33±2.20          & 75.00±4.79          & 75.00±2.98          & 74.99±5.28          & 74.06±4.34           & \textBF{77.94±3.55}  \\
vehicle       & 61.22±4.64          & 60.64±3.67          & 59.10±3.50          & 62.41±3.61          & 62.05±2.76          & \textBF{64.19±4.62}  & 63.37±5.10           \\
vowel         & 63.64±3.43          & 62.22±4.86          & 60.30±5.13          & \textBF{65.15±4.32} & 57.88±3.40          & 63.43±4.51           & 64.04±4.62           \\
wine          & 98.30±2.74          & 97.75±3.92          & \textBF{98.86±2.41} & 95.98±7.79          & 96.63±4.70          & 92.12±7.24           & 98.30±2.74           \\
wisconsin     & 96.71±1.91          & 96.71±1.91          & 97.28±2.18          & 97.14±2.13          & 97.28±2.18          & 95.13±2.16           & \textBF{97.36±2.37}  \\
yeast         & 56.95±3.05          & 57.96±4.19          & 56.95±3.25          & 56.20±3.97          & 55.19±3.87          & 57.35±3.96           & \textBF{58.83±3.50}  \\ \midrule
AVG           & 74.93               & 76.34               & 74.94               & 76.84               & 76.78               & 77.47                & \textBF{79.16}      \\ \bottomrule
\end{tabular}
}
\label{exp-nb}
\end{table}

As shown in Table~\ref{exp-nb}, the naive Bayes classifier using the proposed MRmD discretization scheme obtains the highest classification accuracy on 24 datasets. Compared with the previous filter-based approaches, Ameva~\cite{gonzalez2009ameva}, CAIM~\cite{kurgan2004caim}, MDLP~\cite{fayyad1993multi}, Modified Chi2~\cite{tay2002modified} and PKID~\cite{yang2001proportional}, the proposed MRmD discretization obtains an average improvement of 4.23\%, 2.82\%, 4.22\%, 2.32\% and 2.38\%, respectively. As a filter-based method, the proposed MRmD outperforms the previously best discretization method, the wrapper-based algorithm, EMD \cite{ramirez2015multivariate}, with an average improvement of 1.69\% over the 45 datasets. 
The performance improvements on some datasets are significant. For example, the classification results of the proposed MRmD on ``auto", ``bands" and ``movement" are more than 8\% higher than EMD \cite{ramirez2015multivariate}. Both ``auto" and ``movement" have a relatively small number of samples but a relatively large number of attributes. The NB classifier easily overfits to these two datasets. By maximizing the discriminant information and the generalization performance at the same time, the proposed MRmD achieves a much better generalization performance than EMD \cite{ramirez2015multivariate} that greedily maximizes the discriminant power for a small number of training samples. 

\begin{table}[!ht]

\caption{\footnotesize{Ranks of the Wilcoxon test when comparing various discretization methods under naive Bayes classification framework. Large rank values in the first row and small rank values in the first column indicate that the proposed MRmD significantly outperforms all the compared discretization methods.}}
\centering
\resizebox{0.9\columnwidth}{!}{
\begin{tabular}{cccccccc}

\toprule
\textbf{Algorithm} & \textbf{MRmD} & \textbf{Ameva} & \textbf{CAIM} & \textbf{MDLP} & \textbf{Modified Chi2} & \textbf{PKID} & \textbf{EMD} \\ \midrule
MRmD               & -             & 1030.5         & 1024.5        & 1007.5        & 1014.0                 & 1009.0        & 789.0        \\
Ameva              & 4.5           & -              & 546.5         & 606.5         & 499.0                  & 545.0         & 348.0        \\
CAIM               & 10.5          & 488.5          & -             & 654.0         & 485.0                  & 522.0         & 337.5        \\
MDLP               & 27.5          & 428.5          & 381.0         & -             & 456.0                  & 470.5         & 303.5        \\
Modified Chi2      & 21.0          & 536.0          & 550.0         & 579.0         & -                      & 522.5         & 358.5        \\
PKID               & 26.0          & 490.0          & 513.0         & 564.5         & 512.5                  & -             & 341.0        \\
EMD                & 246.0         & 687.0          & 697.5         & 731.5         & 676.5                  & 694.0         & -      \\
\bottomrule
\end{tabular}
}
\label{MRmD-sig}
\end{table}

To evaluate the significance of the performance gains, we apply the Wilcoxon signed-rank test~\cite{demvsar2006statistical} to thoroughly compare each pair of algorithms. 
The Wilcoxon signed-rank test is a non-parametric statistical test, which ranks the performance of any two algorithms for each dataset, and compares the ranks for their differences. Table~\ref{MRmD-sig} presents the detailed ranks computed by the Wilcoxon test using naive Bayes classifier. Each entry ${R}_{i,j}$ in Table~\ref{MRmD-sig} is the sum of ranks for all datasets on which the algorithm in the $i$-th row is compared with the algorithm in the $j$-th column. 
For the confidence level of $\alpha = 0.05$ and $N=45$, ${R}_{i,j} > 692$ 
indicates that the algorithm in the $i$-th row is significantly better than the algorithm in the $j$-th column. 
As shown in Table~\ref{MRmD-sig}, the proposed MRmD discretization significantly better than Ameva (${R}_{1,2} = 1030.5$), CAIM (${R}_{1,3} = 1024.5$), MDLP (${R}_{1,4} = 1007.5$), Modified Chi2 (${R}_{1,5} = 1014$), PKID (${R}_{1,6} = 1009$) and EMD (${R}_{1,7} = 789$). These results clearly demonstrate that the proposed MRmD significantly outperforms all the compared discretization methods.

\begin{table}[!ht]
\tiny
\caption{\footnotesize{{Comparisons of different discretization methods under the C4.5 classification framework. The proposed MRmD achieves the best average classification accuracy, and outperforms the second-best method, EMD \cite{ramirez2015multivariate}, by 1.45\% on average.}}}
\centering
\resizebox{0.99\columnwidth}{!}{
\begin{tabular}{cccccccc}
              \toprule
              & \multicolumn{1}{c}{Ameva \cite{gonzalez2009ameva}} & \multicolumn{1}{c}{CAIM \cite{kurgan2004caim}} & \multicolumn{1}{c}{MDLP \cite{fayyad1993multi}} & \multicolumn{1}{c}{Modified Chi2~\cite{tay2002modified}} & \multicolumn{1}{c}{PKID \cite{yang2001proportional}} & \multicolumn{1}{c}{EMD \cite{ramirez2015multivariate}} & \multicolumn{1}{c}{MRmD}                 \\ \midrule
abalone       & 21.49±2.13           & 24.32±1.79          & 25.37±2.25           & 17.49±2.55          & 23.07±2.16  & 23.43±2.15          & \textBF{25.63±1.66}  \\
appendicitis  & 83.36±12.55          & 83.36±12.55         & 83.36±10.99          & 78.55±12.64         & 80.18±2.77  & 87.00±7.15          & \textBF{94.36±6.43}  \\
australian    & 86.67±3.47           & 87.25±4.09          & 86.38±3.36           & 85.80±4.20          & 84.93±3.36  & 85.36±4.27          & \textBF{88.55±4.32}  \\
autos         & 75.49±5.66           & 72.63±8.64          & 76.91±10.29          & \textBF{78.97±9.92} & 76.70±10.52 & 76.44±7.39          & 78.52±6.29           \\
balance       & 74.56±4.55           & 74.72±4.58          & 69.92±5.52           & 66.40±5.75          & 64.82±5.35  & \textBF{80.47±3.95} & 77.13±3.58           \\
banana        & 72.49±2.06           & 63.87±1.58          & 74.85±2.20           & 63.92±1.80          & 70.43±1.84  & 87.30±1.55          & \textBF{87.66±1.25}  \\
bands         & 67.00±5.62           & 64.58±4.89          & 53.78±9.16           & 66.42±4.23          & 61.97±7.01  & 64.94±6.61          & \textBF{74.58±4.79}  \\
banknote      & 90.96±2.09           & 88.92±2.04          & 94.46±1.88           & 95.26±1.81          & 84.84±2.33  & 96.57±1.46          & \textBF{98.32±1.18}  \\
bupa          & 68.07±5.37           & 60.65±9.17          & 57.15±7.40           & 57.04±6.45          & 57.89±3.33  & 68.14±7.84          & \textBF{73.03±3.69}  \\
cleveland     & 55.74±7.41           & 54.84±7.07          & 53.77±5.33           & 54.71±9.70          & 53.45±5.35  & 54.80±6.52          & \textBF{57.12±4.16}  \\
climate       & 92.78±2.68           & 92.78±2.55          & 93.33±2.64           & 91.67±1.49          & 91.48±0.91  & 93.33±2.06          & \textBF{95.74±2.49}  \\
contraceptive & 49.09±3.62           & 51.05±3.16          & 50.45±2.69           & 50.45±4.73          & 48.75±2.99  & 52.61±3.70          & \textBF{53.98±3.10}  \\
crx           & 85.51±5.20           & 87.39±3.87          & 86.81±3.89           & \textBF{87.68±4.70} & 85.22±4.92  & 86.38±3.44          & 86.81±3.58           \\
dermatology   & 95.35±3.86           & 93.18±5.58          & 95.89±2.97           & 95.89±2.97          & 94.54±5.76  & 94.82±3.30          & \textBF{96.45±3.00}  \\
ecoli         & 70.82±5.58           & 74.69±4.94          & 77.71±5.86           & 73.81±4.14          & 66.03±6.90  & 74.72±7.72          & \textBF{79.17±6.71}  \\
flare         & 67.82±4.28           & 67.82±4.28          & 67.54±3.80           & 67.54±3.80          & 67.54±3.80  & 67.26±4.09          & \textBF{67.92±4.61}  \\
glass         & 53.05±5.13           & 67.61±11.90         & \textBF{75.79±10.53} & 62.80±13.03         & 57.88±11.59 & 73.70±9.88          & 74.24±7.77           \\
haberman      & 74.13±6.07           & \textBF{75.12±6.01} & 72.53±3.38           & 73.53±1.00          & 73.53±1.00  & 74.15±3.65          & 74.81±4.54           \\
hayes         & 80.20±7.16           & \textBF{80.20±7.16} & 52.02±8.12           & 72.01±13.25         & 71.07±12.98 & 74.26±9.54          & 71.88±7.53           \\
heart         & 78.89±9.57           & 78.52±11.29         & 79.63±9.44           & 78.89±9.57          & 79.26±7.45  & \textBF{82.22±7.18} & 80.37±4.98           \\
hepatitis     & 78.21±11.80          & 82.71±8.27          & 83.25±7.63           & 80.08±8.97          & 82.58±6.82  & 80.71±6.95          & \textBF{83.92±6.40}  \\
iris          & 93.33±3.14           & 93.33±3.14          & 93.33±3.14           & 93.33±4.44          & 92.67±6.63  & \textBF{95.33±4.27} & 94.67±7.77           \\
mammographic  & 81.59±4.78           & 82.84±5.10          & \textBF{83.15±5.29}           & 82.00±4.41          & 81.17±5.25  & 83.15±5.36          & 82.83±2.76  \\
movement      & 43.33±8.30           & 47.50±7.69          & 60.56±10.29          & 63.06±8.18          & 32.22±10.33 & 63.61±5.19          & \textBF{64.17±9.42}  \\
newthyroid    & 92.53±5.08           & 93.51±5.02          & 94.44±4.21           & 93.98±4.43          & 93.98±3.09  & \textBF{94.94±4.33}          & 94.87±4.90  \\
pageblocks    & 96.56±0.52           & 96.18±0.50          & 96.84±0.60           & 94.74±1.07          & 94.96±0.66  & 96.93±0.65          & \textBF{97.17±0.50}  \\
penbased      & 92.93±0.58           & 88.77±1.35          & 88.66±1.29           & 89.04±1.19          & 67.00±0.61  & \textBF{94.91±0.64} & 94.76±0.73           \\
phoneme       & 78.77±1.96           & 79.13±1.81          & 81.24±2.25           & 75.33±1.40          & 76.81±1.84  & \textBF{84.47±1.74} & 83.85±1.59           \\
pima          & \textBF{74.48±3.27}  & 73.45±5.28          & 73.44±4.25           & 72.03±4.68          & 73.07±5.89  & 74.35±1.75          & 74.36±3.24           \\
saheart       & 69.91±4.99           & 70.55±3.94          & 68.17±5.55           & 69.71±4.62          & 65.80±1.36  & 68.39±5.13          & \textBF{71.66±4.27}  \\
satimage      & 25.07±0.55           & 85.41±1.30          & 84.54±1.55           & 83.87±1.39          & 80.20±0.91  & 84.83±1.25          & \textBF{86.01±1.32}  \\
segment       & 95.37±1.12           & 94.68±1.31          & 93.85±1.45           & 88.31±2.23          & 84.76±1.79  & 96.06±0.76          & \textBF{96.06±1.22}  \\
seismic       & 93.42±0.01           & 93.42±0.01          & 93.42±0.01           & 93.42±0.01          & 93.42±0.01  & 93.34±0.24          & \textBF{93.42±0.01}  \\
sonar         & \textBF{79.69±11.88} & 74.00±6.63          & 76.38±11.94          & 73.98±11.05         & 69.62±10.68 & 77.31±9.65          & 77.86±3.96           \\
spambase      & 93.54±1.20           & \textBF{93.56±1.28} & 92.73±1.30           & 87.64±1.62          & 88.69±1.38  & 92.52±1.13          & 92.89±1.07           \\
specfheart    & 80.50±6.80           & 77.85±6.91          & 72.68±9.79           & 77.52±6.67          & 79.42±1.75  & 82.04±2.11          & \textBF{82.05±4.55}  \\
tae           & 44.54±19.82          & 45.83±16.39         & 34.42±2.36           & 52.96±12.40         & 47.08±12.93 & 53.17±12.30         & \textBF{58.17±12.88} \\
thoracic      & 84.68±0.85           & 84.68±0.85          & 84.68±0.85           & 85.11±0.00          & 85.11±0.00  & 84.68±0.85          & \textBF{85.11±0.00}  \\
titanic       & 77.33±3.04           & 77.74±3.15          & 77.15±2.90           & 77.60±2.96          & 78.92±2.31  & \textBF{79.06±2.21} & 77.60±2.42           \\
transfusion   & 79.27±4.67           & 77.27±2.48          & 76.21±0.41           & 76.21±0.41          & 76.21±0.41  & 77.28±4.54          & \textBF{79.96±4.00}  \\
vehicle       & 67.73±4.17           & 67.39±4.15          & 68.32±5.09           & 68.31±5.59          & 64.54±4.47  & 68.31±3.67          & \textBF{70.10±2.33}  \\
vowel         & 72.83±5.68           & 69.60±3.35          & \textBF{73.23±6.43}  & 71.21±5.52          & 48.48±4.29  & 70.71±5.03          & 72.72±4.78           \\
wine          & \textBF{93.82±4.86}  & 91.01±5.53          & 89.84±7.98           & 92.68±6.48          & 79.74±11.09 & 92.12±7.24          & 93.79±5.98           \\
wisconsin     & 93.71±2.04           & 93.85±1.92          & 94.42±2.81           & 94.71±3.09          & 93.84±2.72  & 94.99±1.48          & \textBF{95.60±3.09}  \\
yeast         & 54.99±3.30           & 53.04±4.20          & 57.22±3.16           & 44.34±3.06          & 38.62±3.41  & 52.43±3.49          & \textBF{58.89±4.08}  \\ \hline
AVG           & 75.15                & 76.24               & 76.00                & 75.56               & 72.50       & 78.52               & \textBF{79.97} \\ \bottomrule
\end{tabular}
}
\label{exp-c4.5}
\end{table}


{The proposed MRmD discretization method can be used to boost the performance of not only naive Bayes, but also many other classifiers such as decision tree. We hence include decision tree (C4.5)~\cite{quinlan2014c4} in the comparison, and summarize the results in Table~\ref{exp-c4.5}. The proposed MRmD discretizer obtains the highest classification accuracy on 27 datasets. 
Compared with the previous filter-based methods, Ameva~\cite{gonzalez2009ameva}, CAIM~\cite{kurgan2004caim}, MDLP~\cite{fayyad1993multi}, Modified Chi2~\cite{tay2002modified} and PKID~\cite{yang2001proportional}, the proposed MRmD discretization achieves an average improvement of 4.82\%, 3.73\%, 3.97\%, 4.41\% and 7.47\%, respectively. Compared with the previous best discretization method, EMD~\cite{ramirez2015multivariate}, the proposed MRmD achieves an average improvement of 1.45\% over the 45 datasets. Similarly, we conduct the Wilcoxon signed-rank test~\cite{demvsar2006statistical} on each pair of algorithms to evaluate the significance of the performance gains.  
As shown in Table~\ref{MRmD-sig-c4.5}, the proposed MRmD discretization using C4.5 significantly better than Ameva (${R}_{1,2} = 955.5$), CAIM (${R}_{1,3} = 970$), MDLP (${R}_{1,4} = 998$), Modified Chi2 (${R}_{1,5} = 1014$), PKID (${R}_{1,6} = 1024.5$) and EMD (${R}_{1,7} = 858$).}

\begin{table}[!ht]
\caption{Ranks of the Wilcoxon test when comparing discretization methods using C4.5.}
\centering
\resizebox{0.9\columnwidth}{!}{
\begin{tabular}{cccccccc}
\toprule
\textbf{Algorithm} & \textbf{MRmD} & \textbf{Ameva} & \textbf{CAIM} & \textbf{MDLP} & \textbf{Modified Chi2} & \textbf{PKID} & \textbf{EMD} \\ \midrule
MRmD               & -             & 955.5          & 970.0         & 998.0         & 1014.0                 & 1024.5        & 858.0        \\
Ameva              & 79.5          & -              & 555.0         & 508.5         & 640.5                  & 825.5         & 290.5        \\
CAIM               & 65.0          & 480.0          & -             & 500.5         & 665.5                  & 890.0         & 226.5        \\
MDLP               & 37.0          & 526.5          & 534.5         & -             & 635.5                  & 831.0         & 266.5        \\
Modified Chi2      & 21.0          & 394.5          & 369.5         & 399.5         & -                      & 771.5         & 157.5        \\
PKID               & 10.5          & 209.5          & 145.0         & 204.0         & 263.5                  & -             & 33.0         \\
EMD                & 177.0         & 744.5          & 808.5         & 768.5         & 877.5                  & 1002.0        & -            \\
\bottomrule
\end{tabular}
}
\label{MRmD-sig-c4.5}
\end{table}

\begin{table}[!hpbt]
\centering
\caption{\footnotesize{{Comparisons with the state-of-the-art classifiers.
The proposed MRmD-RNB significantly outperforms the previous best naive Bayes method, RNB \cite{RNB2020shihe}, by 2.84\% on average. Compared with the best deep-learning method, FTT \cite{gorishniy2021revisiting}, the proposed MRmD-RNB obtains an improvement of 0.93\% on average.}}}
    \resizebox{.95\columnwidth}{!}{
\begin{tabular}{cccccccccc}
\toprule
              & \multicolumn{1}{c}{WANBIA~\cite{zaidi2013alleviating}} & \multicolumn{1}{c}{CAWNB~\cite{jiang2019class}} & \multicolumn{1}{c}{AIWNB$^L$~\cite{zhang2021attribute}} & \multicolumn{1}{c}{AIWNB$^E$~\cite{zhang2021attribute}} & \multicolumn{1}{c}{RNB~\cite{RNB2020shihe}} & ResNet \cite{he2016deep}      & PWedRVFL \cite{shi2022weighting}            & FTT \cite{gorishniy2021revisiting}& \multicolumn{1}{c}{MRmD-RNB} \\ \hline
abalone       & 26.71±1.51          & 25.15±1.85  & 26.09±1.44  & 24.01±1.30          & 26.78±1.63  & 25.01±6.25  & 26.35±1.40          & 26.98±2.09          & \textBF{27.00±1.90}  \\
appendicitis  & 87.55±9.07          & 87.55±9.07  & 84.91±9.07  & 84.91±9.07          & 87.55±9.07  & 86.91±8.85  & 82.91±10.47         & 85.82±6.27          & \textBF{88.64±8.49}  \\
australian    & 86.81±3.19          & 86.81±4.64  & 85.35±4.61  & 84.77±4.65          & 86.80±4.34  & 67.32±5.95  & 79.99±4.82          & 86.66±2.94          & \textBF{86.94±5.14}  \\
auto          & 75.62±5.41          & 76.15±11.05 & 71.26±8.47  & 71.31±9.16          & 81.13±9.13  & 65.38±9.95  & 30.57±10.00         & 75.78±7.15          & \textBF{84.35±9.31}  \\
balance       & 71.86±3.89          & 71.86±3.89  & 70.08±2.91  & 71.53±3.14          & 71.86±3.89  & 88.79±3.93  & 87.68±1.24          & 89.93±4.48          & \textBF{91.04±1.70}  \\
banana        & 72.83±2.31          & 73.38±2.04  & 73.32±2.00  & 71.98±2.49          & 73.38±2.04  & 73.72±4.06  & 87.25±2.13          & \textBF{87.42±1.79} & 73.36±1.88           \\
bands         & 70.49±5.86          & 70.69±6.84  & 70.12±6.23  & 70.12±6.23          & 70.69±6.84  & 66.60±7.04  & 64.57±4.35          & 69.01±6.72          & \textBF{75.31±3.89}  \\
banknote      & 92.13±1.40          & 92.78±1.83  & 92.57±1.48  & 92.06±1.38          & 92.78±1.83  & 91.70±17.28 & \textBF{99.93±0.22} & 93.81±2.26          & 92.86±1.87           \\
bupa          & 53.27±10.03         & 53.27±10.03 & 42.02±0.89  & 42.02±0.89          & 53.27±10.03 & 70.09±9.85  & 68.13±8.41          & \textBF{70.73±6.81} & 70.49±8.43           \\
clevland      & 57.73±5.62          & 58.45±4.74  & 58.15±4.65  & 57.17±5.59          & 58.57±7.37  & 58.40±4.31  & 58.76±5.91          & 56.55±6.37          & \textBF{59.06±6.35}  \\
climate       & 94.26±2.66          & 94.26±2.66  & 94.26±2.66  & 94.26±2.66          & 94.26±2.66  & 91.48±0.96  & 91.13±2.23          & 93.87±2.65          & \textBF{94.45±4.16}  \\
contraceptive & 51.38±4.72          & 51.79±4.33  & 51.12±4.16  & 50.72±3.98          & 52.34±4.58  & 52.01±4.82  & 52.21±4.61          & \textBF{53.57±1.78} & 53.43±2.73  \\
crx           & \textBF{87.11±5.59} & 86.67±5.29  & 85.94±5.17  & 85.21±5.26          & 86.38±4.97  & 78.41±11.13 & 80.73±4.47          & 86.67±4.74          & 86.82±5.20  \\
dermatology   & 98.64±1.92          & 98.37±2.28  & 97.56±2.00  & 97.56±2.00          & 98.64±1.92  & 97.27±3.62  & 96.17±3.88          & 97.82±2.67          & \textBF{98.65±2.30}  \\
ecoli         & 82.51±4.00          & 83.38±3.58  & 82.23±5.18  & 82.23±5.18          & 83.39±3.53  & 83.96±6.34  & 84.03±5.22          & 83.03±5.54          & \textBF{84.59±5.59}  \\
flare-solar   & 68.01±4.74          & 68.01±4.74  & 68.20±4.99  & 68.20±4.99          & 68.20±4.99  & 65.87±3.73  & 67.64±4.37          & 67.82±5.64          & \textBF{68.29±5.43}  \\
glass         & 71.13±8.74          & 72.47±5.30  & 75.28±8.17  & 74.28±7.23          & 71.97±4.72  & 56.10±9.21  & 64.61±7.21          & 70.08±8.97          & \textBF{75.71±8.38}  \\
haberman      & 73.18±3.90          & 73.18±3.90  & 26.47±0.72  & 26.47±0.72          & 73.18±3.90  & 73.21±5.01  & 72.81±8.46          & 68.94±6.89          & \textBF{74.80±4.29}  \\
hayes         & 60.03±1.42          & 60.03±1.42  & 60.03±1.42  & 60.03±1.42          & 60.03±1.42  & 77.50±15.65 & 66.28±15.60         & 78.17±9.62          & \textBF{80.09±11.34} \\
heart         & 85.19±8.90          & 85.56±8.27  & 83.70±9.91  & 83.70±9.91          & 85.19±8.90  & 81.85±8.09  & 78.89±11.60         & 81.85±10.66         & \textBF{85.93±9.04}  \\
hepatitis     & 82.17±11.98         & 83.42±9.34  & 82.13±13.16 & 82.79±12.83         & 84.04±9.83  & 85.75±8.51  & 83.42±8.40          & 84.63±10.94         & \textBF{87.25±8.85}  \\
iris          & 93.33±6.29          & 93.33±6.29  & 92.67±6.63  & 92.67±6.63          & 93.33±6.29  & 95.33±6.33  & \textBF{96.00±3.27} & 96.00±4.42          & 94.00±4.92           \\
mammographic  & 82.52±4.20          & 82.52±4.35  & 82.32±4.00  & 82.42±3.83          & 82.63±4.38  & 81.27±4.04  & 81.29±3.98          & 82.31±3.74          & \textBF{83.36±4.94}  \\
movement      & 67.16±3.63          & 68.45±6.00  & 67.12±4.56  & 64.63±5.09          & 68.75±8.39  & 71.95±9.75  & \textBF{75.15±7.47} & 71.14±6.84          & 71.81±8.97           \\
newthyroid    & 95.80±3.50          & 95.35±3.18  & 95.32±4.98  & 95.76±4.73          & 95.80±3.50  & 94.91±4.56  & 89.35±3.51          & 95.78±5.40          & \textBF{98.57±3.21}  \\
pageblocks    & 96.04±0.87          & 96.40±0.70  & 93.79±1.13  & 93.02±1.37          & 96.35±0.97  & 93.48±1.65  & 94.72±0.62          & 96.24±0.77          & \textBF{96.46±0.78}  \\
penbased      & 89.88±0.68          & 92.91±0.75  & 93.60±0.67  & 88.82±0.95          & 93.12±0.72  & 94.08±1.72  & \textBF{95.84±1.35} & 94.00±0.73          & 93.70±0.69  \\
phoneme       & 80.27±1.74          & 80.14±1.87  & 79.79±1.38  & 78.13±1.36          & 79.94±1.76  & 79.29±2.74  & 82.59±1.78          & \textBF{82.68±1.36} & 80.50±2.25  \\
pima          & 74.21±5.02          & 75.12±5.88  & 73.82±4.90  & 73.69±4.58          & 74.86±5.70  & 74.47±2.92  & 75.24±5.51          & 74.61±5.04          & \textBF{75.90±3.56}  \\
saheart       & 70.35±3.06          & 69.91±3.93  & 67.74±5.10  & 67.74±5.10          & 70.12±4.70  & 71.44±5.89  & 72.97±4.16          & 68.83±5.62          & \textBF{73.38±4.84}  \\
satimage      & 84.40±1.08          & 84.27±1.22  & 85.44±0.87  & 81.40±1.43          & 85.86±0.86  & 82.18±1.41  & 85.16±1.28          & 84.99±0.54          & \textBF{86.12±1.14}  \\
segment       & 94.72±1.22 & 93.77±1.23  & 94.20±1.76  & 92.64±1.81          & 94.50±1.02  & 92.77±2.87  & 90.22±1.51          & \textBF{95.11±1.57} & 94.59±1.12           \\
seismic       & 93.42±0.01          & 93.42±0.01  & 82.66±3.04  & 81.19±2.97          & 93.42±0.01  & 93.38±0.13  & 89.86±1.77          & 93.03±0.81          & \textBF{93.46±0.12}  \\
sonar         & 78.37±9.17          & 76.99±10.02 & 76.97±9.42  & 76.49±9.84          & 77.42±9.46  & 77.93±10.01 & 76.53±6.63          & \textBF{81.14±9.98} & 80.28±10.73          \\
spambase      & 93.78±0.97          & 94.07±0.75  & 90.16±1.33  & 89.94±1.32          & 93.98±1.29  & 91.11±1.19  & 89.83±1.93          & 93.57±1.01          & \textBF{94.46±0.69}  \\
specfheart    & 78.54±8.72          & 78.56±7.28  & 75.07±11.65 & 75.07±11.65         & 81.15±9.48  & 78.26±5.54  & 76.36±5.48          & 78.21±6.00          & \textBF{82.70±7.77}  \\
tae           & 34.40±1.79          & 34.40±1.79  & 32.44±1.61  & 32.44±1.61          & 34.40±1.79  & 47.13±9.51  & 47.60±15.25         & 52.92±12.41         & \textBF{58.57±15.67} \\
thoracic      & 83.83±1.49          & 83.40±1.96  & 82.34±2.25  & 81.91±3.05          & 83.83±1.79  & 84.90±0.67  & \textBF{85.53±0.85} & 84.26±1.70          & 84.26±1.79           \\
titanic       & 77.60±2.40          & 77.60±2.40  & 77.60±2.40  & 77.60±2.40          & 77.60±2.40  & 77.88±1.48  & \textBF{79.10±1.00} & 79.05±1.53          & 77.74±2.43           \\
transfusion   & 76.21±0.43          & 76.21±0.43  & 74.47±4.40  & 74.47±4.40          & 76.21±0.43  & 76.21±0.43  & 77.80±4.58          & \textBF{79.68±2.37} & 77.80±3.91           \\
vehicle       & 65.62±5.28          & 65.61±3.74  & 64.66±4.05  & 61.36±6.21          & 67.73±3.21  & 69.03±5.88  & \textBF{78.16±6.32} & 74.70±2.99          & 69.87±5.40           \\
vowel         & 64.14±5.50          & 64.55±4.72  & 66.87±4.99  & 63.64±4.69 & 64.65±5.28  & 64.15±3.76  & \textBF{71.52±5.47} & 68.69±4.52          & 65.15±5.50           \\
wine          & 98.30±2.74          & 97.19±3.96  & 96.60±2.93  & 97.71±2.96          & 98.30±2.74  & 93.33±7.31  & 96.01±3.71          & 96.63±2.75          & \textBF{98.30±2.74}  \\
wisconsin     & 96.93±2.61          & 97.22±2.32  & 97.07±2.75  & 97.36±2.37          & 97.22±2.78  & 96.64±2.07  & 95.61±2.27          & 95.47±2.71          & \textBF{97.36±2.16}  \\
yeast         & 56.75±4.29          & 57.42±4.32  & 57.15±4.00  & 57.15±3.64          & 57.35±4.19  & 55.59±4.70  & 55.47±3.02          & 56.74±3.22          & \textBF{59.30±3.06}  \\ \midrule
AVG           & 77.23               & 77.38       & 75.13       & 74.50               & 77.75       & 77.20       & 77.38               & 79.66               & \textBF{80.59}      \\ \bottomrule
\end{tabular}
}
\label{exp-res2}
\end{table}

\begin{table}[!ht]
\large
\caption{\footnotesize{Ranks of the Wilcoxon test when comparing state-of-the-art classifiers.}}
\centering
\resizebox{0.9\columnwidth}{!}{
\begin{tabular}{@{}cccccccccc@{}}
\toprule
\textbf{Algorithm} & \textbf{MRmD-RNB} & \textbf{WANBIA} & \textbf{CAWNB} & \textbf{AIWNB$^L$} & \textbf{AIWNB$^E$} & \textbf{RNB} & \textbf{ResNet} & \textbf{PWedRVFL} & \textbf{FTT} \\ \midrule
MRmD-RNB           & -                 & 1019.5          & 1034.0         & 1015.0            & 1034.0            & 1029.5       & 998.5           & 788.0             & 751.0        \\
WANBIA             & 15.5              & -               & 408.0          & 817.0             & 946.0             & 232.0        & 593.5           & 496.0             & 334.0        \\
CAWNB              & 1.0               & 627.0           & -              & 868.0             & 973.0             & 262.5        & 621.5           & 507.0             & 358.0        \\
AIWNB$^L$           & 20.0              & 218.0           & 167.0          & -                 & 780.5             & 94.0         & 450.0           & 376.0             & 145.0        \\
AIWNB$^E$           & 1.0               & 89.0            & 62.0           & 254.5             & -                 & 42.5         & 341.0           & 337.5             & 115.5        \\
RNB                & 5.5               & 803.0           & 772.5          & 941.0             & 992.5             & -            & 676.0           & 543.0             & 433.5        \\
ResNet             & 36.5              & 441.5           & 413.5          & 585.0             & 694.0             & 359.0        & -               & 448.0             & 182.0        \\
PWedRVFL           & 247.0             & 539.0           & 528.0          & 659.0             & 697.5             & 492.0        & 587.0           & -                 & 284.0        \\
FTT                & 284.0             & 701.0           & 677.0          & 890.0             & 919.5             & 591.5        & 853.0           & 751.0             & -                 \\ \bottomrule
\end{tabular}
}
\label{RNB-sig}
\end{table}

\subsection{Comparisons to state-of-the-art classifiers}
The proposed MRmD is integrated with RNB~\cite{RNB2020shihe}, named MRmD-RNB, and compared with five state-of-the-art NB classifiers including WANBIA~\cite{zaidi2013alleviating}, CAWNB~\cite{jiang2019class}, AIWNB$^L$~\cite{zhang2021attribute}, AIWNB$^E$~\cite{zhang2021attribute} and RNB~\cite{RNB2020shihe}, and three deep-learning models including ResNet \cite{he2016deep}, FTT \cite{gorishniy2021revisiting} and PWedRVFL \cite{shi2022weighting}. 

As shown in Table~\ref{exp-res2}, the proposed MRmD-RNB obtains the highest classification accuracy on 29 datasets out of 45 datasets among all the compared methods. Compared with the recent attribute weighting methods, AIWNB$^L$ and AIWNB$^E$~\cite{zhang2021attribute}, the proposed MRmD-RNB achieves an average improvement of 5.46\% and 6.09\%, respectively. Compared with the previous best naive Bayes classifier, RNB \cite{RNB2020shihe}, the proposed MRmD-RNB achieves an average improvement of 2.84\%. 
The performance gains on some of the datasets are significant. For example, the performance gains on ``balance", ``bupa", ``hayes" and ``tae" are more than 17\% over RNB~\cite{RNB2020shihe}. Among them, both ``hayes" and ``tae" have a relatively small number of instances and attributes. The NB classifier may overfit to these small datasets due to few training samples. 
The proposed discretization method greatly enhances the generalization ability of the state-of-the-art NB classifier and hence significantly improves the classification performance.

{The proposed MRmD-RNB achieves a higher average classification accuracy than the three compared deep-learning models. 
We conjecture that due to the lack of sufficient representative training samples, the deep-learning models may overfit to the training data. The proposed MRmD-RNB well boosts the generalization capability in both discretization and classifier design, and hence demonstrates excellent classification performance.}

We also conduct the Wilcoxon signed-rank test on the performance gains over the state-of-the-art classifiers. As shown in Table~\ref{RNB-sig}, the proposed MRmD-RNB significantly outperforms all the compared methods, as all the ranks in the first row are much larger than the significance value of 692.

\subsection{{Ablation study}}
{To evaluate the performance gain brought by the generalization capability, we compare the proposed method with the following discretization methods under the naive Bayes classification framework.}

\noindent {\textbf{MR$^O$}: Only the Max-Relevance criterion is used. This serves as the baseline which does not consider the generalization capability at all.}

\noindent {\textbf{MR$^C$}: Only the Max-Relevance criterion is used, but the number of cut points is restricted to the number of classes, as in CAIM~\cite{kurgan2004caim} and  CACC~\cite{tsai2008discretization}.}

\noindent {\textbf{MR$^T$}: Only the Max-Relevance criterion is used, but the number of cut points is restricted to the twice number of cut points derived by MRmD.}

{The average accuracy over 10-fold cross-validation and the average number of cut points are summarized in Table~\ref{exp-ablation}. The proposed MRmD obtains the highest  classification accuracy averaged over 45 datasets. 
Compared with \textbf{MR$^O$} that does not consider the generalization capability, the proposed MRmD achieves an improvement of 5.72\% on average, which is the performance gain brought by the generalization capability through the proposed MRmD discretization scheme. 
By restricting the number of cut points to the number of classes, \textbf{MR$^C$} enhances the generalization capability, but it may lead to a severe loss in discriminant information, as the number of classes may be as small as 2. The proposed MRmD hence outperforms \textbf{MR$^C$} by 3.77\% on average. \textbf{MR$^T$} utilizes twice as many cut points as MRmD, which leads to a decrease of 2.81\% on average from MRmD. This set of results demonstrate that the proposed MRmD can better balance the discriminant power and generalization capability, thus achieving higher classification accuracy.}
\begin{table}[!ht]
\tiny
\caption{\footnotesize{{Comparisons of different discretization methods when constraining the number of cut points using: 1) the number of classes, \textbf{MR$^C$}; 2) MRmD criterion, \textbf{MRmD}; 3) twice the number of cut points derived by MRmD, \textbf{MR$^T$}; 4) no constraint at all, \textbf{MR$^O$}.}}}
\centering
\resizebox{0.99\columnwidth}{!}{
\begin{tabular}{ccccccccc}
\toprule
              & \multicolumn{2}{c}{\textbf{MR$^C$}}         & \multicolumn{2}{c}{\textbf{MRmD}}             & \multicolumn{2}{c}{\textbf{MR$^{T}$}} & \multicolumn{2}{c}{\textbf{MR$^O$}}                 \\ \hline
              & Acc                  & \# of Cut Points     & Acc                  & \# of Cut Points     & Acc                       & \# of Cut Points          & Acc                  & \# of Cut Points       \\ \midrule
abalone       & 25.32±1.64           & 198.00± 0.00  & \textBF{25.54±1.70}  & 71.30±3.89    & 25.46±1.83                & 140.60±7.78        & 19.55±1.76           & 5515.80±12.36   \\
appendicitis  & 85.00±7.82           & 14.00± 0.00   & \textBF{87.91±12.10} & 3.70±1.34     & 85.18±10.52               & 7.40±2.67          & 84.91±7.75           & 138.60±7.50     \\
australian    & 86.22±3.89           & 24.00± 0.00   & 86.37±3.86           & 29.50±4.55    & \textBF{87.09±3.62}       & 55.60±8.93         & 71.44±3.69           & 717.80±7.86     \\
auto          & 65.83±8.00           & 123.80± 0.42  & 76.93±10.75          & 204.30±11.71  & 75.59±9.98                & 382.10±20.55       & \textBF{78.46±10.87} & 759.00±7.47     \\
balance       & 86.40±2.01           & 12.00± 0.00   & 91.04±1.70           & 14.70±1.25    & 78.10±6.45                & 16.00±0.00         & \textBF{91.84±0.48}  & 16.00±0.00      \\
banana        & 58.77±2.02           & 4.00± 0.00    & \textBF{72.96±2.44}  & 16.60±4.14    & 70.43±3.31                & 33.20±8.28         & 60.85±1.83           & 2529.70±11.58   \\
bands         & 66.60±4.01           & 38.00± 0.00   & 73.64±6.42           & 34.10±1.85    & 69.94±6.90                & 67.20±4.13         & \textBF{75.88±3.60}  & 643.90±5.99     \\
banknote      & 85.13±4.33           & 8.00± 0.00    & \textBF{91.55±1.64}  & 28.30±3.50    & 88.19±3.43                & 56.60±7.00         & 82.14±3.17           & 1671.60±14.19   \\
bupa          & 60.27±7.49           & 12.00± 0.00   & \textBF{68.42±7.29}  & 13.40±2.46    & 61.71±10.08               & 26.80±4.92         & 60.57±8.33           & 251.90±5.38     \\
clevland      & 57.36±6.12           & 37.00± 0.00   & 58.07±8.89           & 17.70±2.83    & \textBF{59.12±2.79}       & 3.80±2.04          & 53.24±5.80           & 319.60±3.06     \\
climate       & 91.10±1.73           & 36.00± 0.00   & \textBF{93.51±3.38}  & 12.00±1.83    & 91.66±1.83                & 24.00±3.65         & 91.49±0.86           & 1337.10±15.88   \\
contraceptive & 46.44±2.76           & 21.00± 0.00   & \textBF{52.21±3.02}  & 12.40±1.26    & 46.58±2.87                & 19.80±2.04         & 50.65±2.95           & 61.70±0.48      \\
crx           & 84.93±6.45           & 26.00± 0.00   & \textBF{86.23±6.25}  & 12.80±0.92    & 84.93±6.45                & 22.90±1.79         & 74.77±4.15           & 721.80±5.83     \\
dermatology   & 97.01±2.97           & 89.30± 0.67   & \textBF{98.63±1.95}  & 40.40±0.97    & 95.03±4.11                & 70.00±1.33         & 96.99±3.03           & 140.30±1.06     \\
ecoli         & 82.42±6.18           & 41.90± 0.32   & \textBF{84.28±5.05}  & 15.50±1.35    & 79.98±7.44                & 29.10±2.77         & 68.81±8.30           & 288.90±2.23     \\
flare-solar   & 68.01±5.45           & 12.00± 0.00   & 68.29±5.43           & 11.30±1.42    & 68.29±5.92                & 15.70±1.89         & \textBF{68.66±6.14}  & 17.60±0.70      \\
glass         & 68.15±6.77           & 54.00± 0.00   & \textBF{75.67±9.03}  & 18.30±1.89    & 64.58±8.04                & 36.60±3.78         & 57.91±9.16           & 635.40±10.73    \\
haberman      & 73.58±9.21           & 6.00± 0.00    & \textBF{74.80±4.29}  & 2.90±1.20     & 72.92±6.66                & 5.80±2.39          & 71.59±4.32           & 75.70±2.50      \\
hayes         & \textBF{84.05±10.19} & 11.00± 0.00   & 80.09±11.34          & 7.90±1.10     & 80.44±10.04               & 9.70±1.25          & 83.46±8.16           & 11.00±0.00      \\
heart         & 81.11±9.79           & 22.50± 0.71   & \textBF{84.07±8.91}  & 9.80±1.32     & 81.48±10.33               & 17.80±2.30         & 71.85±8.04           & 265.50±2.95     \\
hepatitis     & 82.63±7.83           & 34.60± 0.97   & \textBF{86.54±11.19} & 39.50±8.63    & 83.25±7.96                & 73.90±16.97        & 81.42±7.81           & 185.20±3.85     \\
iris          & 91.33±5.49           & 12.00± 0.00   & \textBF{94.00±7.34}  & 10.60±1.43    & 90.00±7.86                & 21.20±2.86         & 92.00±8.20           & 52.80±2.04      \\
mammographic  & 80.96±5.21           & 10.00± 0.00   & \textBF{83.36±4.16}  & 19.70±4.81    & 82.62±3.17                & 34.60±8.83         & 82.42±4.14           & 75.90±1.52      \\
movement      & 67.88±9.12           & 1350.00± 0.00 & \textBF{68.90±8.84}  & 1395.90±43.86 & 68.02±6.06                & 2791.80±87.73      & 48.57±8.64           & 16060.10±147.14 \\
newthyroid    & 95.82±4.05           & 15.00± 0.00   & \textBF{97.66±3.96}  & 12.30±1.06    & 95.80±3.44                & 24.60±2.12         & 95.35±5.83           & 133.40±3.89     \\
pageblocks    & 90.55±1.45           & 50.00± 0.00   & \textBF{94.10±0.85}  & 454.50±31.84  & 93.15±1.28                & 909.00±63.68       & 93.75±1.00           & 3147.50±27.28   \\
penbased      & 83.55±0.97           & 160.00± 0.00  & \textBF{88.67±0.86}  & 252.30±5.95   & 86.28±0.84                & 504.60±11.89       & 87.85±0.88           & 1581.40±2.12    \\
phoneme       & 76.26±2.25           & 10.00± 0.00   & \textBF{79.13±2.30}  & 6.00±0.00     & 76.06±2.42                & 12.00±0.00         & 75.83±1.68           & 5589.50±19.03   \\
pima          & 73.03±4.57           & 16.00± 0.00   & \textBF{74.61±3.58}  & 64.00±11.24   & 72.91±4.52                & 128.00±22.49       & 67.57±6.11           & 783.60±9.09     \\
saheart       & 69.91±6.18           & 17.00± 0.00   & \textBF{70.79±3.58}  & 18.30±4.06    & 70.56±6.07                & 35.60±8.11         & 59.07±5.94           & 936.70±8.97     \\
satimage      & 79.81±1.70           & 216.00± 0.00  & 82.28±1.37           & 559.70±9.39   & 82.11±1.42                & 1119.40±18.79      & \textBF{82.39±1.23}  & 2245.80±5.20    \\
segment       & 85.76±2.54           & 116.90± 0.32  & \textBF{92.29±1.69}  & 205.50±8.14   & 89.48±2.05                & 406.70±16.47       & 83.03±1.38           & 9186.50±34.23   \\
seismic       & 85.64±2.60           & 27.00± 0.00   & \textBF{93.42±0.01}  & 0.40±0.70     & 93.42±0.01                & 0.80±1.40          & 91.14±1.67           & 1025.40±8.76    \\
sonar         & 75.49±10.49          & 120.00± 0.00  & \textBF{78.40±10.08} & 43.90±4.28    & 74.99±10.69               & 87.80±8.56         & 63.82±12.68          & 4275.50±39.92   \\
spambase      & 90.98±1.37           & 113.90± 0.32  & 90.53±1.75           & 341.20±31.30  & 90.55±1.44                & 660.30±59.18       & \textBF{91.24±1.12}  & 7480.40±18.82   \\
specfheart    & 75.96±8.83           & 88.00± 0.00   & \textBF{79.71±6.80}  & 172.80±15.53  & 77.83±7.36                & 341.90±27.99       & 79.38±4.87           & 946.40±12.95    \\
tae           & 46.11±12.83          & 11.00± 0.00   & 56.57±15.45          & 55.10±6.64    & \textBF{58.58±16.50}      & 78.40±3.50         & 57.46±10.58          & 83.00±1.41      \\
thoracic      & 81.91±2.70           & 22.00± 0.00   & \textBF{82.98±3.47}  & 9.40±2.07     & 82.55±2.62                & 14.00±3.27         & 81.28±4.11           & 193.10±3.28     \\
titanic       & 77.96±2.24           & 4.00± 0.00    & \textBF{78.19±2.34}  & 2.90±0.32     & 77.96±2.24                & 3.90±0.32          & 77.87±2.35           & 5.00±0.00       \\
transfusion   & 75.14±4.74           & 8.00± 0.00    & \textBF{77.94±3.55}  & 18.50±3.54    & 76.33±3.02                & 37.00±7.07         & 73.53±4.46           & 139.40±4.17     \\
vehicle       & 58.16±5.82           & 72.00± 0.00   & \textBF{63.37±5.10}  & 178.60±13.23  & 61.10±4.61                & 353.20±23.86       & 61.58±4.80           & 1174.40±4.53    \\
vowel         & 56.67±6.52           & 123.00± 0.00  & \textBF{64.04±4.62}  & 85.40±5.02    & 54.75±5.34                & 169.70±9.31        & 23.03±3.77           & 6479.50±14.74   \\
wine          & 94.87±6.85           & 39.00± 0.00   & \textBF{98.30±2.74}  & 16.70±0.48    & 94.31±8.58                & 33.40±0.97         & 92.16±4.74           & 655.60±7.89     \\
wisconsin     & 96.78±2.64           & 18.00± 0.00   & \textBF{97.36±2.37}  & 18.00±1.25    & 96.63±2.40                & 36.00±2.49         & 97.36±2.37           & 70.50±1.27      \\
yeast         & 45.85±6.05           & 63.00± 0.00   & \textBF{58.83±3.50}  & 17.10±1.10    & 39.98±3.76                & 33.20±2.20         & 50.47±3.60           & 368.00±2.75     \\ \midrule
AVG           & 75.39                & 77.93         & \textBF{79.16}       & 101.67        & 76.35                     & 198.93             & 73.44                & 1755.41        \\ \bottomrule
\end{tabular}
}
\label{exp-ablation}
\end{table}

\section{Conclusion}
\label{sec:conclu}
Previous data discretization methods often overemphasize maximizing the discriminant information while overlooking the primary goal of data discretization in classification, \ie, to enhance the generalization ability of a classifier. To address this problem, a Maximal-Dependency-Minimal-Divergence scheme is proposed to simultaneously maximize the generalization capability and discriminant information. The proposed MDmD criterion is difficult to implement in practice due to the difficulty in estimating the high-order mutual information. We hence propose a more practical solution, Maximal-Relevance-Minimal-Divergence criterion, which discretizes one attribute at a time in a top-down manner. The proposed MRmD criterion generates a discretization scheme with a trade-off between retaining the discriminant information and improving the generalization ability for the subsequent classifier. Experimental results on the 45 benchmark datasets demonstrate that the proposed MRmD significantly outperforms all the compared discretization methods and, by integrating the proposed MRmD with RNB, the resulting MRmD-RNB significantly outperforms all the compared classifiers.

{The performance gain of the proposed MRmD may be limited by two factors: 1) The greedy top-down hierarchical splitting algorithm only leads to a near-optimal discretization scheme. Other more sophisticated search algorithms such as genetic algorithm and hyperheuristics can be used to approximate the optimal solution better. 2) The proposed MRmD simplifies the MDmD criterion by ignoring the high-order feature interaction as it is difficult to reliably estimate the multivariate distributions, which may lead to some information loss. A possible improvement is to consider the high-order feature interaction when designing the discretization criterion.}


\section*{Acknowledgment}
This work was supported in part by the National Natural Science Foundation of China under Grant 72071116, and in part by the Ningbo Municipal Bureau Science and Technology under Grants 2019B10026.




{\footnotesize
\bibliographystyle{jabbrv_elsarticle-num}
\bibliography{Discretization-NB}
}

\end{document}